\pdfoutput=1
\documentclass{article}

\usepackage{amsmath,amsfonts,bm}

\def\eqref#1{equation~\ref{#1}}

\def\1{\bm{1}}

\def\vtheta{{\bm{\theta}}}
\def\va{{\bm{a}}}
\def\vb{{\bm{b}}}
\def\vc{{\bm{c}}}

\def\ve{{\bm{e}}}
\def\vf{{\bm{f}}}
\def\vg{{\bm{g}}}
\def\vh{{\bm{h}}}

\def\vu{{\bm{u}}}
\def\vv{{\bm{v}}}

\def\vx{{\bm{x}}}
\def\vy{{\bm{y}}}
\def\vz{{\bm{z}}}

\def\mA{{\bm{A}}}
\def\mB{{\bm{B}}}

\def\mD{{\bm{D}}}
\def\mE{{\bm{E}}}

\def\mI{{\bm{I}}}
\def\mJ{{\bm{J}}}

\def\mU{{\bm{U}}}
\def\mV{{\bm{V}}}

\DeclareMathAlphabet{\mathsfit}{\encodingdefault}{\sfdefault}{m}{sl}
\SetMathAlphabet{\mathsfit}{bold}{\encodingdefault}{\sfdefault}{bx}{n}

\def\gF{{\mathcal{F}}}
\def\gG{{\mathcal{G}}}

\def\gL{{\mathcal{L}}}
\def\gM{{\mathcal{M}}}

\def\gO{{\mathcal{O}}}
\def\gP{{\mathcal{P}}}

\def\gR{{\mathcal{R}}}

\newcommand{\E}{\mathbb{E}}

\newcommand{\R}{\mathbb{R}}

\usepackage[nonatbib,final]{neurips_2021}

\usepackage[utf8]{inputenc} 
\usepackage[T1]{fontenc}    
\usepackage{url}            
\usepackage{booktabs}       
\usepackage{amsfonts}       
\usepackage{nicefrac}       
\usepackage{microtype}      
\usepackage{overpic}        
\usepackage{algorithm}
\usepackage{algpseudocode}
\usepackage{listings}
\usepackage{wrapfig}
\usepackage{amsthm}
\usepackage{colortbl}
\usepackage{makecell}
\usepackage[dvipsnames]{xcolor}
\usepackage[pagebackref=true,breaklinks=true,colorlinks,bookmarks=false,citecolor=blue,linkcolor=blue]{hyperref}

\definecolor{codeblue}{rgb}{0.25, 0.5, 0.5}
\definecolor{codekw}{rgb}{0.35, 0.35, 0.75}
\lstdefinestyle{Pytorch}{
    language         = Python,
    backgroundcolor  = \color{white},
    basicstyle       = \ttfamily\selectfont,
    columns          = fullflexible,
    breaklines       = true,
    captionpos       = b,
    commentstyle     = \fontsize{8pt}{8pt}\color{codeblue},
    keywordstyle     = \fontsize{8pt}{8pt}\color{codekw},
    morekeywords     = with,
}

\graphicspath{{./fig/}}
\DeclareGraphicsExtensions{.pdf,.jpg,.png}

\newtheorem{thm}{Theorem}

\usepackage{cleveref}
\crefname{equation}{Eq.}{Eq.}
\crefname{figure}{Fig.}{Fig.}
\crefname{table}{Tab.}{Tab.~}
\crefname{section}{Sec.}{Sec.~}
\crefname{algorithm}{Alg.}{Alg.~}
\crefname{thm}{Theorem}{Theorem~}
\crefname{lemma}{Lemma}{Lemma~}
\crefname{appendix}{Appendix}{Appendix~}

\def\ie{\textit{i.e.,~}}
\def\eg{\textit{e.g.,~}}
\def\etc{\textit{etc.~}}
\def\etal{\textit{et al.~}}
\def\wrt{\textit{w.r.t.~}}

\def\sota{state-of-the-art~}

\def\pg{\widehat{\frac{\partial\mathcal{L}}{\partial \vtheta}}}
\def\pgn{\widehat{\frac{\partial\mathcal{L}_{n}}{\partial \vtheta}}}
\def\J{\left( \mI - \frac{\partial \mathcal{F}}{\partial \vh} \right)}
\def\Jinline{\left( \mI - \partial \mathcal{F} / \partial \vh \right)}
\def\invJ{\J^{-1}}
\def\invJinline{\Jinline^{-1}}

\title{On Training Implicit Models}

\author{
    Zhengyang Geng\textsuperscript{1,2}\footnotemark[1]
    \quad
    Xin-Yu Zhang\textsuperscript{2}\footnotemark[1]
    \quad
    Shaojie Bai\textsuperscript{4}
    \quad
    Yisen Wang\textsuperscript{2,3}
    \quad
    Zhouchen Lin\textsuperscript{2,3,5}\footnotemark[2]
    \\\\
    $^1$Zhejiang Lab, China \quad $^2$Key Lab. of Machine Perception, School of AI, Peking University \\
    $^3$Institute for Artificial Intelligence, Peking University \\
    \quad $^4$Carnegie Mellon University \quad $^5$Pazhou Lab, China
}

\begin{document}

\maketitle

\vspace{-0.4cm}
\begin{abstract}
    This paper focuses on training implicit models of infinite layers.
    Specifically, previous works employ implicit differentiation and solve the exact gradient for the backward propagation. 
    However, \emph{is it necessary to compute such an exact but expensive gradient for training?}
    In this work, we propose a novel gradient estimate for implicit models, named \emph{phantom gradient},
    that 1) forgoes the costly computation of the exact gradient; 
    and 2) provides an update direction empirically preferable to the implicit model training.
    We theoretically analyze the condition under which an ascent direction
    of the loss landscape could be found, and provide two specific instantiations of the phantom gradient based on the damped unrolling and Neumann series.
    Experiments on large-scale tasks demonstrate that these lightweight phantom gradients significantly accelerate the backward passes in training implicit models by roughly 1.7$\times$, and even boost the performance over approaches based on the exact gradient on ImageNet.
\end{abstract}

\renewcommand{\thefootnote}{\fnsymbol{footnote}}
\footnotetext[1]{Equal contribution}
\footnotetext[2]{Corresponding author, \href{mailto:zlin@pku.edu.cn}{zlin@pku.edu.cn}}
\renewcommand*{\thefootnote}{\arabic{footnote}}

\section{Introduction}
Conventional neural networks are typically constructed by explicitly stacking multiple linear and non-linear operators in a feed-forward manner.
Recently, the implicitly-defined models \cite{chen2018neural,bai2019deep,bai2020multiscale,gu2020implicit,el2021iDL}
have attracted increasing attentions and are able to match the state-of-the-art results by explicit models
on several vision \cite{bai2020multiscale,wang2020iFPN}, language \cite{bai2019deep} and graph \cite{gu2020implicit} tasks.
These works treat the evolution of the intermediate hidden states as a certain form of dynamics,
such as fixed-point equations \cite{bai2019deep,bai2020multiscale} or ordinary differential equations (ODEs) \cite{chen2018neural,dupont2019augmented},
which represents infinite latent states.
The forward passes of implicit models are therefore formulated as solving the underlying dynamics,
by either black-box ODE solvers \cite{chen2018neural,dupont2019augmented} or root-finding algorithms \cite{bai2019deep,bai2020multiscale}.
As for the backward passes, however, directly differentiating through the forward pass trajectories
could induce a heavy memory overhead \cite{franceschi2017forward,lorraine20optimizing}.
To this end, researchers have developed memory-efficient backpropagation via implicit differentiation,
such as solving a Jacobian-based linear fixed-point equation for the backward pass of deep equilibrium models 
(DEQs) \cite{bai2019deep}, which eventually makes the backpropagation trajectories independent of the forward passes.
This technique allows one to train implicit models with essentially constant memory consumption, as we only need to store the final output and the layer itself without saving any intermediate states.
However, in order to estimate the exact gradient by implicit differentiation,
implicit models have to rely on expensive black-box solvers for backward passes,
\eg ODE solvers or root-solving algorithms. 
These black-box solver usually makes the gradient computation very costly in practice, 
even taking weeks to train \sota implicit models on ImageNet~\cite{ILSVRC15} with 8 GPUs.

This work investigates fast approximate gradients for training implicit models. 
We found that a first-order oracle that produces good gradient estimates is enough
to efficiently and effectively train implicit models, 
circumventing laboriously computing the exact gradient as in prior arts \cite{bai2019deep,bai2020multiscale,gu2020implicit,winston2020monotone,lu2021implicit}.
We develop a framework in which a balanced trade-off is made between the precision
and conditioning of the gradient estimate.
Specifically, we provide the general condition under which the phantom gradient can
provide an ascent direction of the loss landscape.
We further propose two instantiations of phantom gradients in the context of DEQ models,
which are based on the the damped fixed-point unrolling and the Neumann series, respectively.
Importantly, we show that our proposed instantiations satisfy the theoretical condition,
and that the stochastic gradient descent (SGD) algorithm based on the phantom gradient
enjoys a sound convergence property as long as the relevant hyperparameters,
\eg the damping factor, are wisely selected.
Note that our method only affects, and thus accelerates,
the backward formulation of the implicit models,
leaving the forward pass formulation (\ie the root-solving process) and the inference behavior unchanged
so that our method is applicable to a wide range of implicit models, forward solvers, and inference strategies.
We conduct an extensive set of synthetic, ablation, and large-scale experiments
to both analyze the theoretical properties of the phantom gradient and
validate its speedup and performances on various tasks, such as ImageNet \cite{ILSVRC15} classification and Wikitext-103 \cite{WIKI} language modeling. 

Overall, our results suggest that: 1) the phantom gradient estimates an ascent direction;
2) it is applicable to large-scale tasks and is capable of achieving a strong performance
which is comparable with or even better than that of the exact gradient;
and 3) it significantly shortens the total training time needed for implicit models
roughly by a factor of $1.4 \sim 1.7 \times$, and even accelerates the backward passes
by astonishingly $12 \times$ on ImageNet.
We believe that our results provide strong evidence for effectively training implicit models with the lightweight phantom gradient.

\section{Method}

\subsection{Inspection of Implicit Differentiation}
\label{sec:notation}
In this work, we primarily focus on the formulation of implicit models based on root-solving, represented by the DEQ models \cite{bai2019deep}.
The table of notations is arranged in \cref{sec:notation-supp}.
Specifically, given an equilibrium module $\gF$, the output of the implicit model is characterized by the solution $\vh^*$ to the following fixed-point equation:
\begin{equation}
  \vh^* = \gF(\vh^*, \vz),
  \label{eq:equation-to-solve}
\end{equation}
where $\vz \in \R^{d_\vu + d_\vtheta}$ is the union of the module's input
$\vu \in \R^{d_\vu}$ and parameters $\vtheta \in \R^{d_\vtheta}$,
\ie $\vz^\top = [\vu^\top, \vtheta^\top]$.
Here, $\vu$ is usually a projection of the original data point $\vx\in\R^{d_\vx}$,
\eg $\vu = \gM(\vx)$.
In this section, we assume $\gF$ is a contraction mapping \wrt $\vh$ so that its Lipschitz constant $L_\vh$
\wrt $\vh$ is less than one, \ie $L_\vh < 1$, a setting that has been analyzed in recent works \cite{pabbaraju2021estimating,revay2020lipschitz}\footnote{
  Note that the contraction condition could be largely relaxed to the one that
  the spectral radius of $\partial \gF / \partial \vh^*$ on the given data
  is less than 1, \ie $\rho(\partial \gF / \partial \vh^*) < 1$,
  as indicated by the well-posedness condition in \cite{el2021iDL}.
}.

To differentiate through the fixed point by \cref{eq:equation-to-solve},
we need to calculate the gradient of $\vh^{*}$ \wrt the input $\vz$.
By Implicit Function Theorem (IFT), we have
\begin{equation}
    \frac{\partial\vh^*}{\partial\vz} = \left. \frac{\partial\gF}{\partial\vz} \right|_{\vh^*} \left( \mI - \left. \frac{\partial \mathcal{F}}{\partial \vh} \right|_{\vh^{*}} \right)^{-1}.
    \label{eq:imp-diff}
\end{equation}
Here, $\left(\partial \va / \partial \vb \right)_{ij} = \partial \va_j / \partial \vb_i$.
The equilibrium point $\vh^{*}$ of \cref{eq:equation-to-solve} is then passed to a post-processing function $\gG$
to obtain a prediction $\hat{\vy} = \gG(\vh^{*})$.
In the generic learning scenario, the training objective is the following expected loss:
\begin{equation}
  \gR(\vtheta) = \E_{(\vx,\vy) \sim \gP} \left[ \gL(\hat{\vy}(\vx; \vtheta),\vy) \right],
  \label{eq:train-objective}
\end{equation}
where $\vy$ is the groundtruth corresponding to the training example $\vx$,
and $\gP$ is the data distribution.
Here, we omit the parameters of $\gG$, because given the output $\vh^*$ of the implicit module $\gF$,
training the post-processing part $\gG$ is the same as training explicit neural networks.
The most crucial component is the gradient of the loss function $\gL$
\wrt the input vector $\vz^\top = [\vu^\top, \vtheta^\top]$,
which is used to train both the implicit module $\gF$ and the input projection module $\gM$.
Using \cref{eq:imp-diff} with the condition $\vh = \vh^*$, we have
\begin{equation}
    \frac{\partial \gL}{\partial \vu} 
    = \frac{\partial\gF}{\partial\vu} \invJ \frac{\partial \gL}{\partial \vh}, 
    \quad
    \frac{\partial \gL}{\partial \vtheta} 
    = \frac{\partial\gF}{\partial \vtheta} \invJ \frac{\partial \gL}{\partial \vh}.
    \label{eq:true-grad}
\end{equation}
The gradients in \cref{eq:true-grad} are in the same form \wrt $\vu$ and $\vtheta$.
Without loss of generality, we only discuss the gradient \wrt $\vtheta$ in the following sections.

\subsection{Motivation}
\label{sec:motivation}

The most intriguing part lies in the Jacobian-inverse term, \ie $\invJinline$.
Computing the inverse term by brute force is intractable due to the $\mathcal{O}(n^3)$ complexity.
Previous implicit models \cite{bai2019deep} approach this by solving a linear system
involving a Jacobian-vector product iteratively via a gradient solver,
introducing over 30 Broyden \cite{broyden1965class} iterations in the backward pass.
However, the scale of the Jacobian matrix can exceed $10^6 \times 10^6$ in the real scenarios, leading to a prohibitive cost in computing the exact gradient.
For example, training a small-scale state-of-the-art implicit model on ImageNet can consume weeks
using 8 GPUs while training explicit models usually takes days,
demonstrating that pursuing the exact gradient severely slows down
the training process of implicit models compared with explicit models.

Secondly, because of the inversion operation, we cast doubt on the conditioning of the gradient and the stability of training process from the numerical aspect.
The Jacobian-inverse can be numerically unstable when encountering the ill-conditioning issue.
The conditioning problem might further undermine the training stability,
as studied in the recent work \cite{bai2021stabilizing}.

Plus, the inexact gradient \cite{lorraine20optimizing,finn2017model,shaban2019truncated,Behrmann2019InvertibleRN,geng2021is,samy2021fixed} is widely applied in the previous learning protocol,
like linear propagation \cite{LinearBackprop} and synthetic gradient \cite{jaderberg2017decoupled}.
Here, the Jacobian-inverse is used to calculate the exact gradient
which is not always optimal for model training.
Moreover, previous research has used a moderate gradient noise as a regularization approach \cite{gastaldi2017shake},
which has been shown to play a central role in escaping poor local minima
and improving generalization ability \cite{an1996effects,zhu2019the,wu2020noisy}.

The concerns and observations motivate us to rethink the possibility of
replacing the Jacobian-inverse term in the standard implicit differentiation
with a cheaper and more stable counterpart.
We believe that an exact gradient estimate is not always required, especially for a black-box layer like those in the implicit models.
Hence this work designs an inexact, theoretically sound,
and practically efficient gradient for training implicit models under various settings.
We name the proposed gradient estimate as the \textit{phantom gradient}.

Suppose the Jacobian $\partial \vh^{*} / \partial \vtheta$ is replaced with a matrix $\mA$,
and the corresponding phantom gradient is defined as
\begin{equation}
  \pg := \mA \; \frac{\partial \gL}{\partial \vh}.
  \label{eq:approx-grad}
\end{equation}
Next, we give the general condition on $\mA$ so that the phantom gradient can be guaranteed valid for optimization (\cref{sec:conditions}), 
and provide two concrete instantiations of $\mA$ based on either damped fixed-point unrolling or
the Neumann series (\cref{sec:neumann}).
The proofs of all our theoretical results are presented in \cref{sec:proofs}.

\subsection{General Condition on the Phantom Gradient}
\label{sec:conditions}
Previous research on theoretical properties for the inexact gradient include several aspects, such as the gradient direction \cite{samy2021fixed}, the unbiasedness of the estimator \cite{chen2019residual}, and the convergence theory of the stochastic algorithm \cite{shaban2019truncated,zhang2021semi}.
The following theorem formulates a sufficient condition that the phantom gradient
gives an ascent direction of the loss landscape.
\begin{thm}
  Suppose the exact gradient and the phantom gradient are given by \cref{eq:true-grad,eq:approx-grad}, respectively.
  Let $\sigma_{\text{max}}$ and $\sigma_{\text{min}}$ be the maximal and minimal
  singular value of $\partial \gF / \partial \vtheta$.
  If
  \begin{equation}
    \left\| \mA \J - \frac{\partial \gF}{\partial \vtheta} \right\| < \frac{\sigma_{\text{min}}^{2}}{\sigma_{\text{max}}},
    \label{eq:condition-descent}
  \end{equation}
  then the phantom gradient provides an ascent direction of the function $\gL$, \ie
  \begin{equation}
    \left\langle \pg, \frac{\partial \gL}{\partial \vtheta} \right\rangle > 0.
    \label{eq:descent-ppt}
  \end{equation}
  \label{thm:descent-condition}
  \vspace{-10pt}
\end{thm}
\paragraph{Remark 1.}
Suppose only the $\invJinline$ term is replaced with a matrix $\mD$,
namely, $\mA = \left( \partial \gF / \partial \vtheta \right) \mD$.
Then, the condition in (\ref{eq:condition-descent}) can be reduced into
\begin{equation}
  \left\| \mD \J - \mI \right\| < \frac{1}{\kappa^{2}},
  \label{eq:condition-descent-simple}
\end{equation}
where $\kappa$ is the condition number of $\partial \gF / \partial \vtheta$.
(See \cref{sec:proof-thm1} for the derivation.)

\subsection{Instantiations of the Phantom Gradient}
\label{sec:neumann}
In this section, we present two practical instantiations of the phantom gradient.
We also verify that the general condition in \cref{thm:descent-condition}
can be satisfied if the hyperparameters in our instantiations are wisely selected.

Suppose we hope to differentiate through implicit dynamics,
\eg either a root-solving process or an optimization problem.
In the context of hyperparameter optimization (HO), previous solutions include differentiating through the unrolled steps
of the dynamics \cite{shaban2019truncated} or employing the Neumann series of the Jacobian-inverse term \cite{lorraine20optimizing}.
In our case, if we solve the root of \cref{eq:equation-to-solve} via the fixed-point iteration:
\begin{equation}
  \vh_{t+1} = \gF(\vh_{t}, \vz),
  \quad
  t = 0,1,\cdots,T-1,
  \label{eq:exact-fpi}
\end{equation}
then by differentiating through the unrolled steps of \cref{eq:exact-fpi}, we have
\begin{equation}
  \frac{\partial \vh_{T}}{\partial \vtheta}
  = \sum_{t=0}^{T-1} \left. \frac{\partial \gF}{\partial \vtheta} \right|_{\vh_{t}} \prod_{s=t+1}^{T-1} \left. \frac{\partial \gF}{\partial \vh} \right|_{\vh_{s}}.
  \label{eq:exact-bptt}
\end{equation}
Besides, the Neumann series of the Jacobian-inverse $\invJinline$ is
\begin{equation}
  \mI + \frac{\partial \gF}{\partial \vh}
  + \left( \frac{\partial \gF}{\partial \vh} \right)^{2}
  + \left( \frac{\partial \gF}{\partial \vh} \right)^{3}
  + \cdots.
  \label{eq:original-neumann}
\end{equation}
Notably, computing the Jacobian $\partial \vh^{*} / \partial \vtheta$ using the Neumann series in (\ref{eq:original-neumann})
is equivalent to differentiating through the unrolled steps of \cref{eq:exact-fpi} at the exact equilibrium point $\vh^{*}$
and taking the limit of infinite steps \cite{lorraine20optimizing}.

\textit{Without altering the root} of \cref{eq:equation-to-solve}, we consider
a damped variant of the fixed-point iteration:
\begin{equation}
  \vh_{t+1} = \gF_{\lambda}(\vh_{t}, \vz) = \lambda \gF(\vh_{t}, \vz) + (1 - \lambda) \vh_{t},
  \quad
  t = 0,1,\cdots,T-1.
  \label{eq:damped-fpi}
\end{equation}
Differentiating through the unrolled steps of \cref{eq:damped-fpi},
\cref{eq:exact-bptt} is adapted as
\begin{equation}
  \frac{\partial \vh_{T}}{\partial \vtheta}
  = \lambda \sum_{t=0}^{T-1} \left. \frac{\partial \gF}{\partial \vtheta} \right|_{\vh_{t}} \prod_{s=t+1}^{T-1} \left( \left. \lambda \frac{\partial \gF}{\partial \vh} \right|_{\vh_{s}} + \left( 1 - \lambda \right) \mI \right).
  \label{eq:damped-bptt}
\end{equation}
The Neumann series of $\invJinline$ is correspondingly adapted as
\begin{equation}
  \lambda \left( \mI + \mB + \mB^{2} + \mB^{3} + \cdots \right),
  \quad
  \mbox{where}~~\mB = \lambda \frac{\partial \gF}{\partial \vh} + (1 - \lambda) \mI.
  \label{eq:damped-neumann}
\end{equation}
The next theorem shows that under mild conditions, the Jacobian from the damped unrolling in \cref{eq:damped-bptt}
converges to the exact Jacobian and the Neumann series in (\ref{eq:damped-neumann})
converges to the Jacobian-inverse $\invJinline$ as well.
\begin{thm}
  Suppose the Jacobian $\partial \gF / \partial \vh$ is a contraction mapping.
  Then,
  \begin{enumerate}
    \item[(i)] the Neumann series in (\ref{eq:damped-neumann}) converges to the Jacobian-inverse $\invJinline$; and
    \item[(ii)] if the function $\gF$ is continuously differentiable \wrt both $\vh$ and $\vtheta$,
    the sequence in \cref{eq:damped-bptt} converges to the exact Jacobian $\partial \vh^{*} / \partial \vtheta$ as $T \to \infty$, \ie
    \begin{equation}
      \lim_{T \to \infty} \frac{\partial \vh_{T}}{\partial \vtheta}
      = \left. \frac{\partial\gF}{\partial\vtheta} \right|_{\vh^*} \left( \mI - \left. \frac{\partial \mathcal{F}}{\partial \vh} \right|_{\vh^{*}} \right)^{-1}.
      \label{eq:converge-unroll}
    \end{equation}
  \end{enumerate}
  \label{thm:convergence-of-phantom}
\end{thm}
However, as discussed in \cref{sec:motivation}, it is unnecessary to compute the exact gradient
with infinite terms.
In the following context, we introduce two instantiations of the phantom gradient
based on the finite-term truncation of \cref{eq:damped-bptt} or (\ref{eq:damped-neumann}).

\paragraph{Unrolling-based Phantom Gradient (UPG).}
In the unrolling form, the matrix $\mA$ is defined as
\begin{equation}
  \mA_{k,\lambda}^{\text{unr}} = \lambda \sum_{t=0}^{k-1} \left. \frac{\partial \gF}{\partial \vtheta} \right|_{\vh_{t}} \prod_{s=t+1}^{k-1} \left( \left. \lambda \frac{\partial \gF}{\partial \vh} \right|_{\vh_{s}} + \left( 1 - \lambda \right) \mI \right).
  \label{eq:A-unroll}
\end{equation}

\paragraph{Neumann-series-based Phantom Gradient (NPG).}
In the Neumann form, the matrix $\mA$ is defined as
\begin{equation}
  \mA_{k,\lambda}^{\text{neu}} = \lambda \left. \frac{\partial \gF}{\partial \vtheta} \right|_{\vh^{*}} \left( \mI + \mB + \mB^{2} + \cdots + \mB^{k-1} \right),
  \quad
  \mbox{where}~~\mB = \lambda \left. \frac{\partial \gF}{\partial \vh} \right|_{\vh^{*}} + (1 - \lambda) \mI.
  \label{eq:A-neumann}
\end{equation}

Note that both the initial point of the fixed-point iteration (\ie $\vh_{0}$ in \cref{eq:A-unroll}) and the point at which the Neumann series is evaluated (\ie $\vh^{*}$ in (\ref{eq:A-neumann})) are the solution of the root-finding solver. (See \cref{sec:implementation} for implementation of the phantom gradient.)

According to \cref{thm:convergence-of-phantom}, the matrix $\mA$ defined by
either \cref{eq:A-unroll} or (\ref{eq:A-neumann}) converges to the exact Jacobian
$\partial \vh^{*} / \partial \vtheta$ as $k \to \infty$ for any $\lambda \in (0,1]$.
Therefore, by \cref{thm:convergence-of-phantom}, the condition in (\ref{eq:condition-descent})
can be satisfied if a sufficiently large step $k$ is selected, since
\begin{equation}
  \left\| \mA \J - \frac{\partial \gF}{\partial \vtheta} \right\|
  \le (1 + L_{\vh}) \left\| \mA - \frac{\partial \gF}{\partial \vtheta} \invJ \right\|.
\end{equation}

Next, we characterize the impact of the two hyperparameters, \ie $k$ and $\lambda$,
on the precision and conditioning of $\mA$.
Take the NPG (\cref{eq:A-neumann}) as an example.
\begin{enumerate}
  \item[(i)] On the precision of the phantom gradient,
  \begin{itemize}
    \item a large $k$ makes the gradient estimate more accurate,
    as higher-order terms of the Neumann series are included, while
    \item a small $\lambda$ slows down the convergence of the Neumann series
    because the norm $\| \mB \|$ increases as $\lambda$ decreases.
  \end{itemize}
  \item[(ii)] On the conditioning of the phantom gradient,
  \begin{itemize}
    \item a large $k$ impairs the conditioning of $\mA$
    since the condition number of $\mB^{k}$ grows exponentially as $k$ increases, while
    \item a small $\lambda$ helps maintain a small condition number of $\mA$
    because the singular values of $\partial \gF / \partial \vh$ are
    ``smoothed'' by the identity matrix.
  \end{itemize}
\end{enumerate}
In a word, a large $k$ is preferable for a more accurate $\mA$,
while a small $\lambda$ contributes to the well-conditioning of $\mA$.
Practically, these hyperparameters should be selected in consideration
of a balanced trade-off between the precision and conditioning of $\mA$.
See \cref{sec:experiment} for experimental results.

\subsection{Convergence Theory}
\label{sec:convergence}
In this section, we provide the convergence guarantee of the SGD algorithm
using the phantom gradient.
We prove that under mild conditions, if the approximation error of the phantom gradient is sufficiently small,
the SGD algorithm converges to an $\epsilon$-approximate stationary point in the expectation sense.
We will discuss the feasibility of our assumptions in \cref{sec:proof-thm3}.
\begin{thm}
  Suppose the loss function $\gR$ in \cref{eq:train-objective} is $\ell$-smooth,
  lower-bounded, and has bounded gradient almost surely in the training process.
  Besides, assume the gradient in \cref{eq:true-grad} is an
  unbiased estimator of $\nabla \gR(\vtheta)$ with a bounded covariance.
  If the phantom gradient in \cref{eq:approx-grad} is an $\epsilon$-approximation
  to the gradient in \cref{eq:true-grad}, \ie
  \begin{equation}
    \left\| \pg - \frac{\partial \gL}{\partial \vtheta} \right\| \le \epsilon,
    \quad
    \mbox{almost surely},
    \label{eq:eps-approx}
  \end{equation}
  then using \cref{eq:approx-grad} as a stochastic first-order oracle with a step size of
  $\eta_{n} = \gO (1 / \sqrt{n})$ to update $\vtheta$ with gradient descent,
  it follows after $N$ iterations that
  \begin{equation}
    \E \left[ \frac{\sum_{n=1}^{N} \eta_{n} \| \nabla \gR(\vtheta_{n}) \|^2}{\sum_{n=1}^{N} \eta_{n}} \right] \le \gO \left( \epsilon + \frac{\log{N}}{\sqrt{N}} \right).
  \end{equation}
  \label{thm:convergence-SGD}
\end{thm}
\vspace{-.2in}
\paragraph{Remark 2.}
Consider the condition in (\ref{eq:eps-approx}):
\begin{equation}
  \begin{split}
    \left\| \pg - \frac{\partial \gL}{\partial \vtheta} \right\|
    \le \left\| \mA - \frac{\partial \gF}{\partial \vtheta} \invJ \right\|
    \left\| \frac{\partial \gL}{\partial \vh} \right\|.
  \end{split}
\end{equation}
Suppose the gradient $\partial \gL / \partial \vh$ is almost surely bounded.
By \cref{thm:convergence-of-phantom}, the condition in (\ref{eq:eps-approx})
can be guaranteed as long as a sufficiently large $k$ is selected.

\section{Experiments}
\label{sec:experiment}
\vspace{-0.1cm}
In this section, we aim to answer the following questions via empirical results: 
(1) How precise is the phantom gradient?
(2) What is the difference between the unrolling-based and the Neumann-series-based phantom gradients?
(3) How is the phantom gradient influenced by the hyperparameters $k$ and $\lambda$?
(4) How about the computational cost of the phantom gradient compared with implicit differentiation?
(5) Can the phantom gradient work at large-scale settings for various tasks?

We have provided some theoretical analysis and intuitions to (1), (2), and (3) in \cref{sec:neumann}.
Now we answer (1) and (2) and demonstrate the performance curves
under different hyperparameters $k$ and $\lambda$ on CIFAR-10 \cite{CIFAR}.
Besides, we also study other factors that have potential influences
on the training process of the \sota implicit models \cite{bai2019deep,bai2020multiscale} like pretraining.
For (4) and (5), we conduct experiments on large-scale datasets to highlight
the ultra-fast speed and competitive performances, including image classification
on ImageNet \cite{ILSVRC15} and language modeling on Wikitext-103 \cite{WIKI}.

We start by introducing two experiment settings.
We adopt a single-layer neural network with spectral normalization
\cite{takeru2018spectral} as the function $\gF$ and fixed-point iterations as
the equilibrium solver, which is the \textit{synthetic setting}.
Moreover, on the CIFAR-10 dataset, we use the MDEQ-Tiny model \cite{bai2020multiscale} (170K parameters) as the backbone model, denoted as the \textit{ablation setting}.
Additional implementation details and experimental results are presented in \cref{sec:experiment-supp}\footnote{
  All training sources of this work are available at \url{https://github.com/Gsunshine/phantom_grad}.
}.
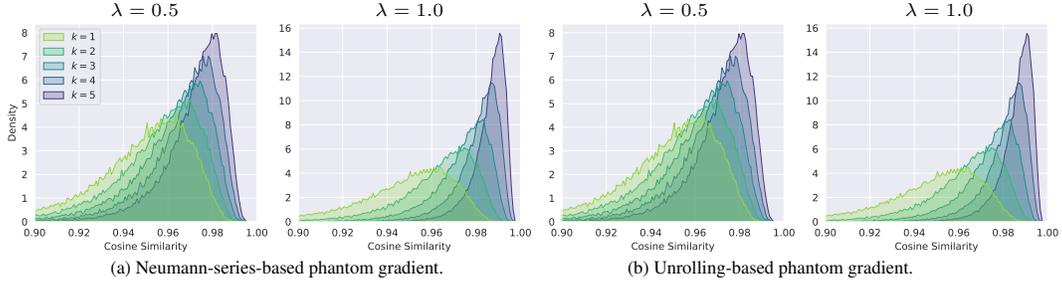
\begin{figure}[!t]
    \centering
    \begin{overpic}[width=\linewidth]{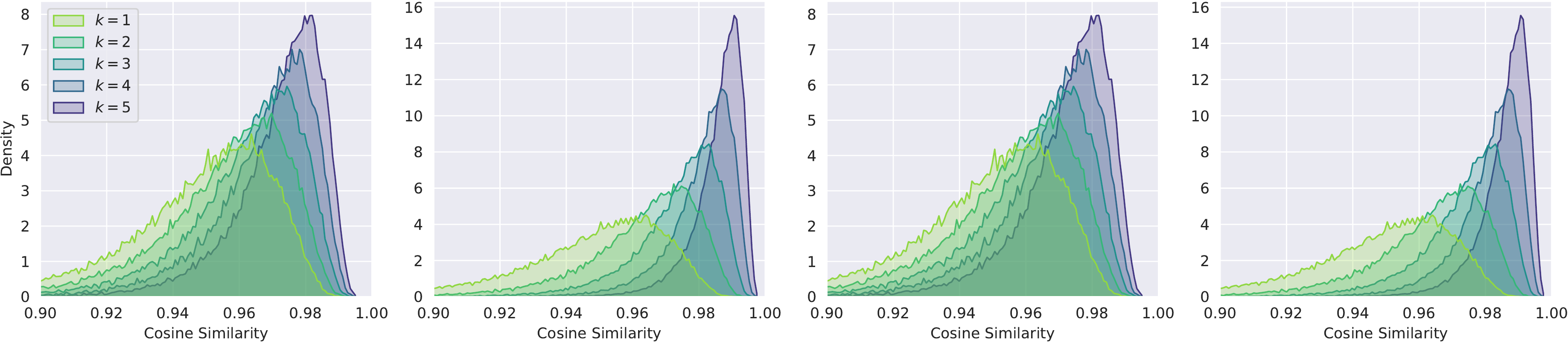}
        \put(9.8,-2.0){\scriptsize{(a) Neumann-series-based phantom gradient.}}
        \put(59.0,-2.0){\scriptsize{(b) Unrolling-based phantom gradient.}}
        \put(9.8,22.5){\scriptsize{$\lambda = 0.5$}}
        \put(35.0,22.5){\scriptsize{$\lambda = 1.0$}}
        \put(60.2,22.5){\scriptsize{$\lambda = 0.5$}}
        \put(85.4,22.5){\scriptsize{$\lambda = 1.0$}}
    \end{overpic}
    \vspace{-1mm}
    \caption{Cosine similarity between the phantom gradient and the exact gradient in the synthetic setting.}
    \label{fig:toy_grad_good}
    \vspace{-3mm}
\end{figure}

\paragraph{Precision of the Phantom Gradient.}
The precision of the phantom gradient is measured by its angle against the exact gradient, indicated by the cosine similarity between the two.
We discuss its precision in both the synthetic setting and the ablation setting.
The former is under the static and randomly generated weights,
while the latter provides characterization of the training dynamics.

In the synthetic setting, the function $\gF$ is restricted to be a contraction mapping.
Specifically, we directly set the Lipschitz constant of $\gF$ as $L_\vh = 0.9$,
and use 100 fixed-point iterations to solve the root $\vh^*$ of \cref{eq:equation-to-solve}
until the relative error satisfies $\| \vh - \gF(\vh, \vz) \| / \| \vh \| < {10}^{-5}$.
Here, the exact gradient is estimated by backpropagation through the fixed-point iterations,
and cross-validated by implicit differentiation solved with 20 iterations
of the Broyden's method \cite{broyden1965class}.
In our experiment, the cosine similarity between these two gradient estimates consistently succeeds
$0.9999$, indicating the gradient estimate is quite accurate when the relative error of forward solver is minor.
The cosine similarity between phantom gradients and exact gradients is shown in \cref{fig:toy_grad_good}.
It shows that the cosine similarity tends to increase as $k$ grows and
that a small $\lambda$ tends to slow down the convergence of the phantom gradient,
allowing it to explore in a wider range regarding the angle against the exact gradient.

\begin{figure}[!ht]
    \centering
    \begin{overpic}[width=0.5\linewidth]{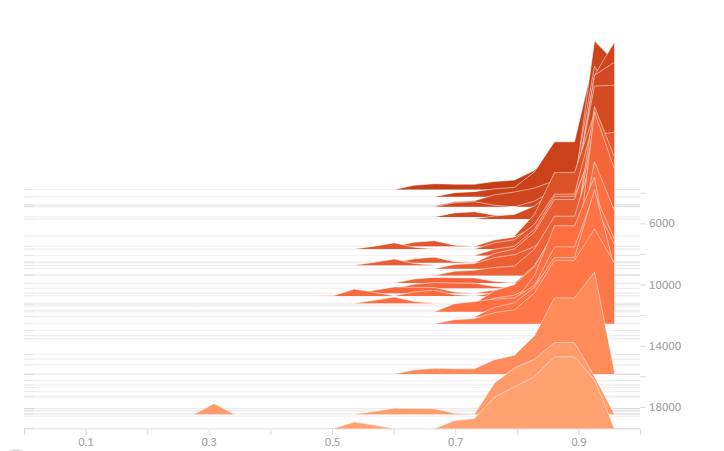}
        \put(14.8,-4.0){\scriptsize{(a) Neumann-series-based phantom gradient.}}
    \end{overpic}%
    \begin{overpic}[width=0.5\linewidth]{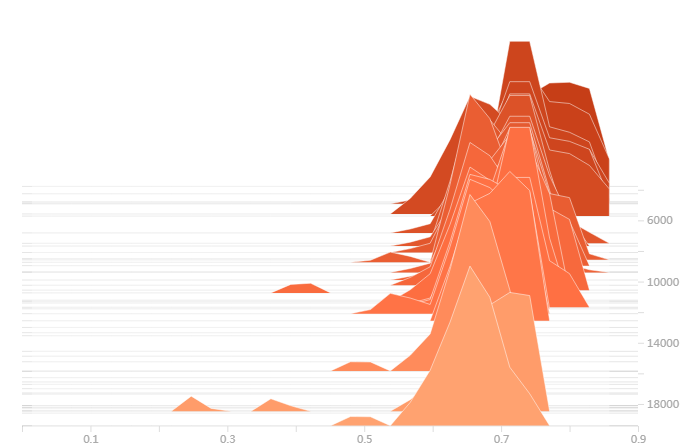}
        \put(24.5,-4.0){\scriptsize{(b) Unrolling-based phantom gradient.}}
    \end{overpic}%
    \vspace{4.5pt}
    \caption{
      Cosine similarity between the phantom gradient and the exact gradient in the real scenario.
      The horizontal axis corresponds to the cosine similarity, and the vertical axis to the training step.
    }
    \label{fig:real_cos}
    \vspace{-5pt}
\end{figure}

In the ablation setting, the precision of the phantom gradient
during the training process is shown in \cref{fig:real_cos}.
The model is trained by implicit differentiation under the official schedule\footnote{Code available at \url{https://github.com/locuslab/mdeq}.}.
It shows that the phantom gradient still provides an ascent direction in the real training process, as indicated by the considerable cosine similarity against the exact gradient.
Interestingly, the cosine similarity slightly decays as the training progresses, which implies a possibility to construct an adaptive gradient solver for implicit models.

\paragraph{To Pretrain, or not to Pretrain?}
To better understand the components of the implicit models' training schedule,
we first illustrate a detailed ablation study of the baseline model in the ablation setting.
The average accuracy with standard deviation is reported in \cref{Tab:ablation}.

\begin{wraptable}{r}{7cm}
\vspace{-0.3cm}
\centering
\makeatletter\def\@captype{table}\makeatother
\caption{Ablation settings on CIFAR-10.}
\vspace{0.1cm}
\begin{tabular}{p{4cm} p{1.5cm}} 
\Xhline{1pt}
\specialrule{0em}{0pt}{1pt}
\textbf{Method} &  \;\textbf{Acc. (\%)} \\
\hline
\specialrule{0em}{0pt}{1pt}
Implicit Differentiation   & $85.0 \pm 0.2$   \\
\hline
\specialrule{0em}{0pt}{1pt}
w/o Pretraining            & $82.3 \pm 1.3$   \\
w/o Dropout                & $83.7 \pm 0.1$   \\
Adam $\rightarrow$ SGD     & $84.5 \pm 0.3$   \\
SGD w/o Pretraining        & $82.9 \pm 1.5$   \\
\hline
\specialrule{0em}{0pt}{1pt}
UPG ($\mA_{5,0.5}$, w/o Dropout)       & $\mathbf{85.8} \pm 0.5$   \\
NPG ($\mA_{5,0.5}$, w/o Dropout)       & $85.6 \pm 0.5$           \\
\hline
\specialrule{0em}{0pt}{1pt}
UPG ($\mA_{9,0.5}$, w/ Dropout)        & $\mathbf{86.1} \pm 0.5$   \\
\Xhline{1pt}
\end{tabular}
\label{Tab:ablation}
\end{wraptable}

The MDEQ model employs a pretraining stage in which the model $\gF$
is unrolled as a recurrent network.
We study the impact of the pretraining stage, the Dropout \cite{hinton2012improving} operation,
and the optimizer separately.
It can be seen that the unrolled pretraining stabilizes the training of the MDEQ model.
Removing the pretraining stage leads to a severe performance drop and apparent training instability
among different trials because the solver cannot obtain an accurate fixed point $\vh^*$
when the model is not adequately trained.
This ablation study also suggests that the MDEQ model is a strong baseline for our method to compare with.

However, pretraining is not always indispensable for training implicit models.
It introduces an extra hyperparameter, \ie how many steps should be involved in the pretraining stage.
Next, we discuss how the UPG could circumvent this issue.

\vspace{-8pt}
\paragraph{Trade-offs between Unrolling and Neumann.}
For an exact fixed point $\vh^*$, \ie $\vh^* = \gF (\vh^*, \vz)$, there is no difference between UPG and NPG.
However, when the numerical error exists in solving $\vh^*$, \ie $\| \vh^{*} - \gF(\vh^{*}, \vz) \| > 0$,
these two instatiations of the phantom gradient can behave differently.

\begin{wraptable}{r}{6.4cm}
  \vspace{-0.4cm}
  \centering
  \makeatletter\def\@captype{table}\makeatother
  \caption{
    Complexity comparison. Mem means the memory cost, and
    $K$ and $k$ denote the solver's steps and the unrolling/Neumann steps, respectively. Here, $K \gg k \approx 1$.
  }
   \vspace{0.15cm}
  \begin{tabular}{p{1.2cm}| p{0.9cm}<{\centering} p{0.9cm}<{\centering} p{1.6cm}<{\centering} } 
    \Xhline{1pt}
    \specialrule{0em}{0pt}{1pt}
    \textbf{Method}  & \textbf{Time} & \textbf{Mem} & \textbf{Peak Mem}\\
    \hline
    \specialrule{0em}{0pt}{1pt}
    Implicit & $\mathcal{O}(K)$   &  $\mathcal{O}(1)$  & $\mathcal{O}(k)$     \\
    UPG      & $\mathcal{O}(k)$   &  $\mathcal{O}(k)$  & $\mathcal{O}(k)$     \\
    NPG      & $\mathcal{O}(k)$   &  $\mathcal{O}(1)$  & $\mathcal{O}(1)$     \\
    \Xhline{1pt}
  \end{tabular}
  \label{Tab:complexity}
\vspace{-0.2cm}
\end{wraptable}

We note that a particular benefit of the UPG is its ability
to automatically switch between the pretraining and training stages for implicit models.
When the model is not sufficiently trained and the solver converges poorly (see \cite{bai2020multiscale}),
the UPG defines a forward computation graph that is essentially equivalent to a shallow weight-tied network to refine the coarse equilibrium states.
In this stage, the phantom gradient serves as a backpropagation through time (BPTT) algorithm
and hence behaves as in the pretraining stage.
Then, as training progresses, the solver becomes more stable and converges to the fixed point $\vh^*$ better.
This makes the UPG behave more like the NPG.
Therefore, the unrolled pretraining is gradually transited into the regular training phase based on
implicit differentiation, and the hyperparameter tuning of pretraining steps can be waived.
We argue that such an ability to adaptively switch training stages is benign to 
the implicit models' training protocol, which is also supported by the performance gain in \cref{Tab:ablation}. 

Although the UPG requires higher memory overhead than implicit differentiation
or the NPG, it does not surpass the peak memory usage in the entire training
protocol by implicit differentiation due to the pretraining stage.
In the ablation setting, the MDEQ model employs a 10-layer unrolling
for pretraining, which actually consumes double memory compared with a 5-step unrolling scheme,
\eg $\mA_{5,0.5}$ in \cref{Tab:ablation}.
In \cref{Tab:complexity}, we also demonstrate the time and memory complexity for
implicit differentiation and the two forms of phantom gradient.

In addition, leaving the context of approximate and exact gradient aside, we also develop insights into understanding the subtle differences between UPG and NPG in terms of the state-dependent and state-free gradient.
Actually, the UPG is state-dependent, which means it corresponds to the ``exact'' gradient of a computational sub-graph.
Both the NPG and the gradient solved by implicit differentiation, however,
do not exactly match the gradient of any forward computation graph
unless the numerical error is entirely eliminated in both the forward
and the backward passes for implicit differentiation or the one-step gradient \cite{wilder2019end,geng2021is,zhang2021semi,samy2021fixed} is used for the NPG, \ie $k=1$.
Interestingly, we observe that models trained on the state-dependent gradient
demonstrate an additional training stability regarding the Jacobian spectral radius,
compared with those trained on the state-free counterpart.
We empirically note that it can be seen as certain form of implicit Jacobian regularization
for implicit models as a supplement to the explicit counterpart \cite{bai2021stabilizing},
indicated by the stable estimated Jacobian spectral radius during training,
\ie $\rho(\partial \gF_{\lambda} / \partial \vh) \approx 1$.

\begin{figure}[!t]
    \centering
    \begin{overpic}[height=5cm]{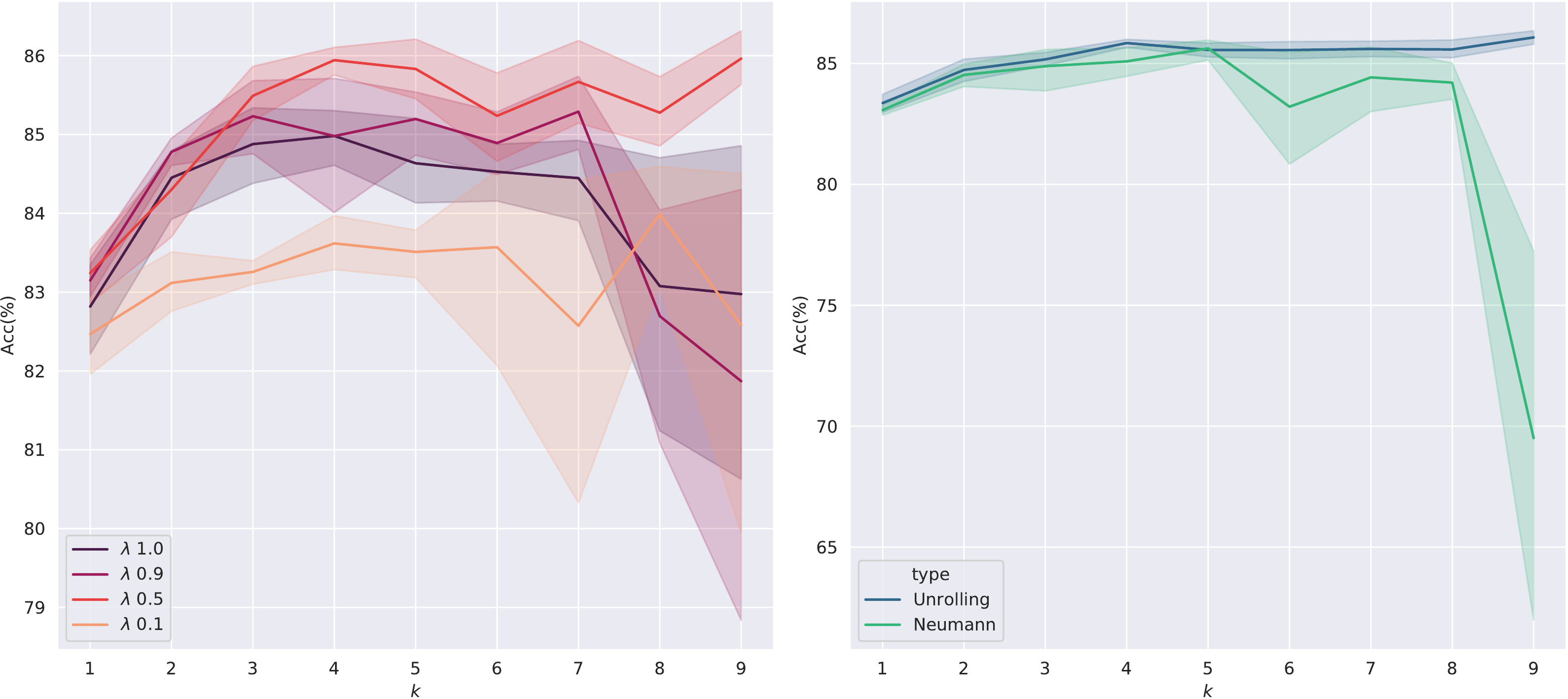}
        \put(17.0,-2.0){\scriptsize{(a) Impact of $\lambda$ and $k$}}
        \put(68.0,-2.0){\scriptsize{(b) NPG \textit{v.s.} UPG}}
    \end{overpic}
    \caption{Ablation studies on (a) the hyperparameters $\lambda$ and $k$,
    and (b) two forms of phantom gradient.}
    \label{fig:exp_neu}
    \vspace{-0.2cm}
\end{figure}

The experimental results also cohere with our insight.
The performance curves in \cref{fig:exp_neu} demonstrate the influence of $\lambda$ and $k$
and further validate that the UPG is more robust to a wide range of steps $k$ than the NPG.
In fact, when the Jacobian spectral radius $\rho(\partial \gF / \partial \vh)$ increases freely without proper regularization,
the exploding high-order terms in the NPG could exert a negative impact on the overall direction of the phantom gradient,
leading to performance degradation when a large $k$ is selected (see \cref{fig:exp_neu}(b)).
We also observe a surging of the estimated Jacobian spectral radius
as well as the gradient explosion issue in the NPG experiments.
On the contrary, the UPG can circumvent these problems thanks to
its implicit Jacobian regularization.

\vspace{-0.1cm}

\begin{table}[ht]
  \centering
  \caption{Experiments using DEQ \cite{bai2019deep} and MDEQ \cite{bai2020multiscale} on vision and language tasks. Metrics stand for accuracy(\%)$\uparrow$ for image classification on CIFAR-10 and ImageNet, and perplexity$\downarrow$ for language modeling on Wikitext-103. JR stands for Jacobian Regularization \cite{bai2021stabilizing}. $\dagger$ indicates additional steps in the forward equilibrium solver.}
  \vspace{-0.1cm}
  \begin{tabular}{lllclc} 
    \Xhline{1pt}
    \specialrule{0em}{0pt}{1pt}
    \textbf{Datasets} & \textbf{Model} & \textbf{Method} & \textbf{Params}  & \multicolumn{1}{c}{\textbf{Metrics}} &  \textbf{Speed} \\ 
    \hline
    \specialrule{0em}{0pt}{1pt}
    CIFAR-10 & MDEQ & Implicit         & 10M & $93.8 \pm 0.17$ & $1.0\times$   \\
    \specialrule{0em}{0pt}{0.5pt}
    CIFAR-10 & MDEQ & UPG $\mA_{5,0.5}$ & 10M & $95.0 \pm 0.16$ & $1.4 \times$ \\
    \hline
    \specialrule{0em}{0pt}{1pt}
    ImageNet & MDEQ & Implicit         & 18M & $75.3$ & $1.0\times$   \\
    \specialrule{0em}{0pt}{0.5pt}
    ImageNet & MDEQ & UPG $\mA_{5,0.6}$ & 18M & $75.7$ & $1.7\times$ \\
    \hline
    \hline
    \specialrule{0em}{0pt}{2pt}
    Wikitext-103 & DEQ (PostLN)  & Implicit         & 98M   & $24.0$ & $1.0\times$   \\
    \specialrule{0em}{0pt}{0.5pt}
    Wikitext-103 & DEQ (PostLN)  & UPG $\mA_{5,0.8}$ & 98M   & $25.7$ & $1.7 \times$  \\
    \hline
    \specialrule{0em}{0pt}{2pt}
    Wikitext-103 & DEQ (PreLN)            & JR + Implicit         & 98M   & $24.5$           & $1.7 \times$ \\
    \specialrule{0em}{0pt}{0.8pt}
    Wikitext-103 & DEQ (PreLN)            & JR + UPG $\mA_{5,0.8}$ & 98M   & $24.4$           & $2.2 \times$ \\
    Wikitext-103 & DEQ (PreLN)            & JR + UPG $\mA_{5,0.8}$ & 98M   & $24.0^\dagger$  & $1.7 \times$ \\
    \Xhline{1pt}
  \end{tabular}
  \label{Tab:deq}
\end{table}

\vspace{-0.15cm}
\paragraph{Phantom Gradient at Scale.}
We conduct large-scale experiments to verify the advantages of the phantom gradient
on vision, graph, and language benchmarks.
We adopt the UPG in the large-scale experiments.
The results are illustrated in \cref{Tab:deq} and \cref{Tab:ignn}.
Our method matches or surpasses the implicit differentiation training protocol
on the \sota implicit models with a visible reduction on the training time. 
When only considering the backward pass, the acceleration for MDEQ can be remarkably $12 \times$ on ImageNet classification. 

\begin{table}[ht]
  \centering
  \caption{Experiments using IGNN \cite{gu2020implicit} on graph tasks. Metrics stand for accuracy(\%)$\uparrow$ for graph classification on COX2 and PROTEINS, Micro-F1(\%)$\uparrow$ for node classification on PPI.}
  \vspace{-0.1cm}
  \begin{tabular}{lllcl}
    \Xhline{1pt}
    \specialrule{0em}{0pt}{1pt}
    \textbf{Datasets} & \textbf{Model} & \textbf{Method} & \textbf{Params}  & \multicolumn{1}{c}{\textbf{Metrics (\%)}}  \\ 
    \hline
    \specialrule{0em}{0pt}{1pt}
    COX2  & IGNN   & Implicit            & 38K & $84.1\pm 2.9$    \\
    COX2  & IGNN   & UPG $\mA_{5,0.5}$    & 38K & $83.9\pm 3.0$    \\
    COX2  & IGNN   & UPG $\mA_{5,0.8}$    & 38K & $83.9\pm 2.7$    \\
    COX2  & IGNN   & UPG $\mA_{5,1.0}$    & 38K & $83.0\pm 2.9$    \\
    \hline
    \specialrule{0em}{0pt}{1pt}
    PROTEINS & IGNN   & Implicit         & 34K & $78.6\pm 4.1$    \\
    PROTEINS & IGNN   & UPG $\mA_{5,0.5}$ & 34K & $78.4\pm 4.2$    \\
    PROTEINS & IGNN   & UPG $\mA_{5,0.8}$ & 34K & $78.6\pm 4.2$    \\
    PROTEINS & IGNN   & UPG $\mA_{5,1.0}$ & 34K & $78.8\pm 4.2$    \\
    \hline
    \specialrule{0em}{0pt}{1pt}
    PPI & IGNN   & Implicit              & 4.7M & $97.6$     \\
    PPI & IGNN   & UPG $\mA_{5,0.5}$      & 4.7M & $98.2$     \\
    PPI & IGNN   & UPG $\mA_{5,0.8}$      & 4.7M & $97.4$     \\
    PPI & IGNN   & UPG $\mA_{5,1.0}$      & 4.7M & $96.2$     \\
    \Xhline{1pt}
  \end{tabular}
  \label{Tab:ignn}
  \vspace{-0.4cm}
\end{table}

\section{Related Work}
\label{sec:related-work}
\vspace{-3pt}
\paragraph{Implicit Models.}
Implicit models generalize the recursive forward/backward rules of neural networks
and characterize their internal mechanism by some pre-specified dynamics.
Based on the dynamics, the implicit mechanisms can be broadly categorized into three classes:
ODE-based \cite{chen2018neural,dupont2019augmented},
root-solving-based \cite{bai2019deep,bai2020multiscale,gu2020implicit,el2021iDL,winston2020monotone},
and optimization-based \cite{amos2017optnet,djolonga2017differentiable,wang2019satnet} implicit models.

The ODE-based implicit models \cite{chen2018neural,dupont2019augmented} treat the iterative update rules
of residual networks as the Euler discretization of an ODE,
which could be solved by any black-box ODE solver.
The gradient of the differential equation is calculated using the \textit{adjoint method} \cite{boltyanskiy1962mathematical},
in which the adjoint state is obtained by solving another ODE.
The root-solving-based implicit models \cite{bai2019deep,el2021iDL,bai2020multiscale,gu2020implicit,wang2020iFPN,winston2020monotone,samy2021fixed}
characterize layers of neural networks by solving fixed-point equations.
The equations are solved by either the black-box root-finding solver
\cite{bai2019deep,bai2020multiscale} or the fixed-point iteration \cite{gu2020implicit,samy2021fixed}.
The optimization-based implicit models \cite{amos2017optnet,djolonga2017differentiable,wang2019satnet,vlastelica2019differentiation,wilder2019end,geng2021is,gould2019deep,xie2021optimization} leverage the optimization programs as layers of neural networks.
Previous works have studied differentiable layers of quadratic programming \cite{amos2017optnet},
submodular optimization \cite{djolonga2017differentiable},
maximum satisfiability (MAXSAT) problems \cite{wang2019satnet},
and structured decomposition \cite{geng2021is}.
As for the backward passes, implicit differentiation is applied to
the problem-defining equations of the root-solving-based models \cite{bai2019deep,bai2020multiscale}
or the KKT conditions of the optimization-based models \cite{amos2017optnet}.
As such, the gradient can be obtained from solving the backward linear system.

In this work, we focus on the root-solving-based implicit models.
Theoretical works towards root-solving-based implicit models include the well-posedness \cite{el2021iDL,gu2020implicit}, monotone operators \cite{winston2020monotone}, global convergence \cite{kawaguchi2020theory,xie2021optimization}, and Lipschitz analysis \cite{revay2020lipschitz}.
We look into the theoretical aspect of the gradient-based algorithm in training implicit models and the efficient practice guidance.
With these considerations, we show that implicit models of the same architecture
could enjoy faster training speed and strong generalization in practical applications by using the phantom gradient.

\vspace{-5pt}
\paragraph{Non-End-to-End Optimization in Deep Learning.}
Non-end-to-end optimization aims to replace the standard gradient-based training of deep architectures with modular or weakly modular training without the entire forward and backward passes.
Currently, there are mainly three research directions in this field, namely,
the auxiliary variable methods \cite{carreira-perpinan2014distributed,taylor2016training,zhang2016efficient,zhang2017convergent,zeng2018global,li2019lifted,gu2020fenchel},
target propagation \cite{bengio2014how,lee2015difference,alexander2020a},
and synthetic gradient \cite{jaderberg2017decoupled,czarnecki2017understanding,lansdell2020learning}.
The auxiliary variable methods \cite{carreira-perpinan2014distributed,taylor2016training,zhang2016efficient,zhang2017convergent,zeng2018global,li2019lifted,gu2020fenchel}
formulate the optimization of neural networks as constrained optimization problems,
in which the layer-wise activations are considered as trainable auxiliary variables.
Then, the equality constraints are relaxed as penalty terms added to the objectives
so that the parameters and auxiliary variables can be divided into blocks and
thus optimized in parallel.
The target propagation method \cite{bengio2014how,lee2015difference,alexander2020a}
trains each module by having its activations regress to the pre-assigned targets,
which are propagated backwards from the downstream modules.
Specifically, the auto-encoder architecture is used to reconstruct targets at each layer.
Finally, the synthetic gradient method \cite{jaderberg2017decoupled,czarnecki2017understanding,lansdell2020learning}
estimates the local gradient of neural networks using auxiliary models,
and employ the synthetic gradient in place of the exact gradient to perform parameter update.
In this way, the forward and backward passes are decoupled and can be executed
in an asynchronous manner.

Our work is in line with the non-end-to-end optimization research
since we also aims to decouple the forward and backward passes of neural networks.
However, we show that finding a reasonable ``target'' or a precise gradient estimate
is not always necessary in training deep architectures.
Our paper paves a path that an inexact but well-conditioned gradient estimate
can contribute to both fast training and competitive generalization of implicit models.

\vspace{-5pt}
\paragraph{Differentiation through Implicit Dynamics.}
Differentiation through certain implicit dynamics is an important aspect in a wide range of research fields,
including bilevel optimization \cite{shaban2019truncated,lorraine20optimizing},
meta-learning \cite{finn2017model,rajeswaran2019meta,zhang2021semi},
and sensitivity analysis \cite{bonnans2013perturbation}.
Since the gradient usually cannot be computed analytically,
researchers have to implicitly differentiate the dynamics at the converged point.
The formula of the gradient typically contains a term of Jacobian-inverse
(or Hessian-inverse), which is computationally prohibitive for large-scale models.
(See \cref{eq:imp-diff} in our case.)
Herein, several techniques have been developed to approximate the matrix inverse
in the previous literature.

An intuitive solution is to differentiate through the unrolled steps of
a numerical solver of the dynamics \cite{domke2012generic,maclaurin2015gradient,franceschi2017forward}.
In particular, if a single step is unrolled, it reduces to the well-known \textit{one-step gradient}
\cite{luketina2016scalable,finn2017model,liu2018darts,wilder2019end,zhang2021semi,geng2021is,samy2021fixed}, in which the inverse of Jacobian/Hessian is simply approximated by an identity matrix.
On the contrary, unrolling a small number of steps may induce a bias \cite{lorraine20optimizing},
while the memory and computational cost grows linearly as the number of unrolled steps increases.
Towards this issue, Shaban \etal \cite{shaban2019truncated} propose to
truncate the long-term dependencies and differentiate through only the last $L$ steps.
In fact, if the dynamics have converged to a stationary point,
the finite-term truncation in Shaban \etal \cite{shaban2019truncated} is exactly
the Neumann approximation of the Jacobian-inverse with the first $L$ terms.
Based on this, Lorraine \etal \cite{lorraine20optimizing} directly
use the truncated Neumann series as an approximation of the Jacobian-inverse.
Besides the unrolling-based methods, optimization-based approaches
\cite{pedregosa2016hyperparameter,rajeswaran2019meta} have been studied in this field as well.
Since the Jacobian-inverse-vector product can be viewed as solution of a linear system,
algorithms like the conjugate gradient method can be used to solve it.

\vspace{-5pt}
\section{Limitation and Future Work}
\vspace{-4pt}

The main limitation of this work lies in the hyperparameter tuning of the phantom gradient,
especially for the damping factor $\lambda$, which directly controls the gradient's precision
and conditioning, the implicit Jacobian regularization for UPG, the stability for NPG,
and the final generalization behaviors.
However, it has not been a hindrance to the application of phantom gradients
in training implicit models as one can tune the hyperparameter according to
the validation loss in the early training stage.

Regarding future works, we would like to highlight the following aspects:
(1) eliminating the bias of the current phantom gradient,
(2) constructing an adaptive gradient solver for implicit models,
(3) analyzing the damping factor to provide practical guidance,
(4) investigating the implicit Jacobian regularization, and
(5) understanding how different noises in the gradients can impact the training
of implicit models under different loss landscapes.

\vspace{-5pt}
\section{Conclusion}
\vspace{-5pt}
In this work, we explore the possibility of training implicit models via the efficient approximate phantom gradient.
We systematically analyze the general condition of a gradient estimate so that
the implicit model can be guaranteed to converge to an approximate stationary point
of the loss function.
Specifically, we give a sufficient condition under which a first-order oracle
could always find an ascent direction of the loss landscape in the training process.
Moreover, we introduce two instantiations of the proposed phantom gradient,
based on either the damped fixed-point unrolling or the Neumann series.
The proposed method shows a $1.4 \sim 1.7 \times$ acceleration with comparable
or better performances on large-scale benchmarks.
Overall, this paper provides a practical perspective on training implicit models
with theoretical guarantees.

\subsubsection*{Acknowledgments}
Zhouchen Lin was supported by the NSF China (No.s 61625301 and 61731018), NSFC Tianyuan Fund for Mathematics (No. 12026606) and Project 2020BD006 supported by PKU-Baidu Fund.
Yisen Wang was partially supported by the National Natural Science Foundation of China
under Grant 62006153, and Project 2020BD006 supported by PKU-Baidu Fund.
Shaojie Bai was sponsored by a grant from the Bosch Center for Artificial Intelligence. 

\bibliographystyle{unsrt}
\bibliography{citation.bib}

\clearpage
\appendix

\section{Table of Notions}
\label{sec:notation-supp}
The notations in this work are summarized in \cref{tab:notations}.
\begin{table}[!h]
  \definecolor{mygray}{RGB}{220,220,220}
  \renewcommand{\arraystretch}{1.5}
  \centering
  \caption{Table of notations in this work.}
  \vspace{-2pt}
  \begin{tabular}{cl}
      \Xhline{1pt}
      \textbf{Symbol} & \multicolumn{1}{c}{\textbf{Description}} \\
      \hline\hline
      \multicolumn{2}{c}{\textbf{Vectors}}\\
      \rowcolor{mygray} $\vx$ & Input data \\
      $\vu$ & Injection of input $\vx$ \\
      \rowcolor{mygray} $\vh$ & Intermediate feature of input $\vx$ \\
      $\hat{\vy}$ & Prediction of input $\vx$ \\
      \rowcolor{mygray} $\vy$ & Groundtruth of input $\vx$ \\
      $\vtheta$ & Parameter vector of the equilibrium module \\
      \rowcolor{mygray} $\vz$ & A union of $\vu$ and $\vtheta$ \\
      \hline\hline
      \multicolumn{2}{c}{\textbf{Functions}}\\
      \rowcolor{mygray} $\gM(\vx)$ & Preprocessing module, $\gM: \R^{d_{\vx}} \to \R^{d_{\vu}}$ \\
      $\gF(\vh,\,\vz)$ & Equilibrium module, $\gF: \R^{d_{\vh}} \times \R^{d_{\vz}} \to \R^{d_{\vh}}$ \\
      \rowcolor{mygray} $\gG(\vh)$ & Postprocessing module, $\gG: \R^{d_{\vh}} \to \R^{d_{\vy}}$ \\
      $\gR(\vtheta)$ & Loss function, $\gR: \R^{d_{\vtheta}} \to \R$ \\
      \hline\hline
      \multicolumn{2}{c}{\textbf{Equilibrium states $\vh$}}\\
      \rowcolor{mygray} $\vh^{*}$ & The equilibrium point of $\gF$ given $\vz$ \\
      $\vh_{t}$ & The intermediate feature of the $t^{\text{th}}$ unrolled step \\
      \hline\hline
      \multicolumn{2}{c}{\textbf{Gradients \& Jacobians}}\\
      \rowcolor{mygray} $\frac{\partial \gL}{\partial \vtheta}$ & Exact gradient of the loss \wrt the parameters $\vtheta$ \\
      $\widehat{\frac{\partial \gL}{\partial \vtheta}}$
      & Phantom gradient, \ie an approximation to $\frac{\partial \gL}{\partial \vtheta}$ \\ \rowcolor{mygray} $\frac{\partial \va}{\partial \vb}$
      & Gradient of $\va$ \wrt $\vb$, \ie $\left(\frac{\partial \va}{\partial \vb}\right)_{ij} = \frac{\partial \va_j}{\partial \vb_i}$. \\
      $\mA$ & An approximation to $\frac{\partial \gF}{\partial \vtheta} \invJ$ \\
      \rowcolor{mygray} $\mD$ & An approximation to $\invJ$ \\
      \hline\hline
      \multicolumn{2}{c}{\textbf{Scalars}}\\
      \rowcolor{mygray} $\sigma_{\text{max}},\,\sigma_{\text{min}}$ & The maximal/minimal singular value of $\frac{\partial \gF}{\partial \vtheta}$ \\
      $\kappa$ & The condition number of $\frac{\partial \gF}{\partial \vtheta}$ \\
      \rowcolor{mygray} $k,\,\lambda$ & The number of steps and damping factor of phantom gradient \\
      $L_{\vh}$ & The Lipschitz constant of $\gF$ \wrt $\vh$ \\
      \rowcolor{mygray} $d_{\diamond}$ & Dimension of vector $\diamond$ \\
      \hline\hline
      \multicolumn{2}{c}{\textbf{Operators}}\\
      \rowcolor{mygray} $\langle\cdot,\,\cdot\rangle$ & Inner product \\
      $\| \cdot \|$ & Vector norm or operator norm \\
      \rowcolor{mygray} $\rho(\cdot)$ & Spectral radius \\
      \Xhline{1pt}
  \end{tabular}
  \label{tab:notations}
\end{table}

\section{Algorithm of Phantom Gradient}
\label{sec:implementation}
The following PyTorch-style \cite{pythorch} pseudocode describes the implementation of
both the unrolling-based phantom gradient (see \cref{alg:UPG-torch})
and the Neumann-series-based one (see \cref{alg:NPG-torch}).
To implement the phantom gradient with TensorFlow \cite{tensorflow},
replace the $no\_grad$ context manager with the $stop\_gradient$ operator.

The unrolling-based phantom gradient is computed by the automatic differentiation engine,
while the Neumann-series-based phantom gradient is given by \cref{alg:NPG}.
A special reminder is that, for a trained model, removing the unrolling steps
in the test stage will not lead to a performance decay but accelerate the inference instead.
Similarly, increasing the unrolling steps in the test stage can not further improve the performance,
which is validated using MDEQ model on the CIFAR-10 and ImageNet datasets.
This implies that the root-finding solver has fully converged to an equilibrium point for the trained model.

\begin{algorithm}[!ht]
\caption{Unrolling-based phantom gradient, PyTorch-style}
\label{alg:UPG-torch}
\begin{lstlisting}[style=Pytorch,escapeinside={(@}{@)}]
# solver: the solver to find (@$\vh^{*}$@), e.g., the Broyden solver in MDEQ.
# func: the explicit function (@$\gF$@) that defines the implicit model.
# z: the input variables (@$\vz$@) to solve (@$\vh^{*} = \gF(\vh^{*}, \vz)$@)
# h: the solution (@$\vh^{*}$@) of the implicit module.
# k: the unrolling steps (@$k$@).
# lambda_: the damping factor (@$\lambda$@).
# training: a bool variable that indicates the training or inference stage.

# Forward pass (Backward pass is accomplished by automatic differentiation)
def forward(z, k, lambda_, training):
    with torch.no_grad():
        h = solver(func, z)   

    if training:
        for _ in range(k):
            h = (1 - lambda_) * h + lambda_ * func(h, z)    

    return h
\end{lstlisting}
\end{algorithm}

\begin{algorithm}[!ht]
\caption{Neumann-series-based Phantom Gradient, Pytorch-style}
\label{alg:NPG-torch}
\begin{lstlisting}[style=Pytorch,escapeinside={(@}{@)}]
# solver: the solver to find (@$\vh^{*}$@), e.g., the Broyden solver in MDEQ.
# func: the explicit function (@$\gF$@) that defines the implicit model.
# grad(a, b, c): the function to compute the Jacobian-vector product (@$\left( \partial \va / \partial \vb \right) \vc$@)
# z: the input variables (@$\vz$@) to solve (@$\vh^{*} = \gF(\vh^{*}, \vz)$@)
# h: the output (@$\vh^{*}$@) of the implicit module.
# g: the input gradient (@$\partial \gL / \partial \vh$@).
# g_out: the output gradient (@$\partial \gL / \partial \vz$@).
# k: the unrolling steps (@$k$@).
# lambda_: the damping factor (@$\lambda$@).

# Forward pass
def forward(z):
    with torch.no_grad():
        h = solver(func, z)

    return h

# Backward pass
def phantom_grad(g, h, z, k, lambda_):
    f = (1 - lambda_) * h + lambda_ * func(h, z)
    
    g_hat = g
    for _ in range(k-1):
        g_hat = g + grad(f, h, g_hat)
        
    g_out = lambda_ * grad(f, z, g_hat)
    return g_out
\end{lstlisting}
\end{algorithm}

\begin{algorithm}[!ht]
\caption{Neumann-series-based phantom gradient with $\gO(1)$ memory}
\label{alg:NPG}
\begin{algorithmic}[1]
    \State Input $\partial \gL / \partial \vh$, $\gF$, $\vh^*$, $k$, $\lambda$.
    \State Initialize $\hat{\vg} = \vg = \partial \gL / \partial \vh$;
    \State $\vf \leftarrow (1 - \lambda) \vh^* + \lambda \gF(\vh^*, \vz)$
    \For{$i=1,2,\cdots,k-1$}
        \State $\hat{\vg} \leftarrow \vg + \left( \partial \vf / \partial \vh \right) \hat{\vg}$; \algorithmiccomment{Compute Jacobian-vector product with automatic differentiation} 
    \EndFor
    \State $\vg_{\text{out}} \leftarrow \lambda \left( \partial \vf / \partial \vz \right) \hat{\vg} $ \algorithmiccomment{Compute Jacobian-vector product to obtain the phantom gradient \wrt $\vz$}
    \State \Return $\hat{\vg}$.
\end{algorithmic}
\end{algorithm}

\section{Proof of Theorems}
\label{sec:proofs}
\subsection{Proof of \cref{thm:descent-condition}}
\label{sec:proof-thm1}
\begin{proof}
  Denote $\mJ = \partial \gF / \partial \vtheta$, $\vv = \partial \gL / \partial \vh$,
  and $\vu = \left( \mI - \partial \gF / \partial \vh \right)^{-1} \vv$.
  Let
  \begin{equation}
    \mE = \mA \J - \frac{\partial \gF}{\partial \vtheta},
  \end{equation}
  and we have $\left\| \mE \right\| < \sigma_{\text{min}}^{2} / \sigma_{\text{max}}$.
  Then,
  \begin{equation}
    \begin{split}
      \left\langle \pg, \frac{\partial \gL}{\partial \vtheta} \right\rangle
      &= \vv^{\top} \mA^{\top} \mJ \left( \mI - \frac{\partial \gF}{\partial \vh} \right)^{-1} \vv
      = \vu^{\top} \left( \mI - \frac{\partial \gF}{\partial \vh} \right)^{\top} \mA^{\top} \mJ \vu
      = \vu^{\top} \left( \mJ + \mE \right)^{\top} \mJ \vu
      \\
      &\ge \left\| \mJ \vu \right\|^{2} - \| \mE \| \| \mJ \| \| \vu \|^{2}
      \ge \left( \sigma_{\text{min}}^{2} - \sigma_{\text{max}} \| \mE \| \right)
      \| \vu \|^{2}
      > 0,
    \end{split}
  \end{equation}
  which concludes the proof.
\end{proof}
\begin{proof}[Proof of Remerk 1]
  Suppose $\mA = \left( \partial \gF / \partial \vtheta \right) \mD$ and
  the condition in (\ref{eq:condition-descent-simple}).
  Then,
  \begin{equation}
    \left\| \mA \J - \frac{\partial \gF}{\partial \vtheta} \right\|
    \le \left\| \frac{\partial \gF}{\partial \vtheta} \right\| \left\| \mD \J - \mI \right\|
    < \sigma_{\text{max}} \cdot \frac{1}{\kappa^{2}}
    = \frac{\sigma_{\text{min}}^{2}}{\sigma_{\text{max}}},
  \end{equation}
  indicating the condition in (\ref{eq:condition-descent}) is satisfied.
\end{proof}
\subsection{Proof of \cref{thm:convergence-of-phantom}}
\begin{proof}
  (i) Since $\left\| \partial \gF / \partial \vh \right\| < 1$,
  \begin{equation}
    \left\| \mB \right\|
    \le \lambda \left\| \frac{\partial \gF}{\partial \vh} \right\| + (1 - \lambda) \left\| \mI \right\|
    < 1.
    \label{eq:contractive-B}
  \end{equation}
  Let $\mB_{k} = \sum_{t=0}^{k-1} \mB^{t}$, and for each $p \in \mathbb{N}_{+}$, we have
  \begin{equation}
    \left\| \mB_{k+p} - \mB_{k} \right\| = \left\| \sum_{t=k}^{k+p-1} \mB^{t} \right\|
    \le \left\| \mB \right\|^{k} \left\| \sum_{t=0}^{p-1} \mB^{t} \right\|
    \le \left\| \mB \right\|^{k} \sum_{t=0}^{p-1} \left\| \mB \right\|^{t}
    < \frac{\left\| \mB \right\|^{k}}{1 - \left\| \mB \right\|}.
  \end{equation}
  By the Cauchy's convergence test, the sequence $\{ \mB_{k} \}$ is convergent.
  Since
  \begin{equation}
    (\mI - \mB) \mB_{k} = \mI - \mB^{k} \to \mI,
    \quad
    \mbox{as}~k \to \infty,
  \end{equation}
  it follows that $\mB_{k} \to \left( \mI - \mB \right)^{-1}$, as $k \to \infty$.
  Therefore,
  \begin{equation}
    \lambda \sum_{t=0}^{\infty} \mB^{t} = \lambda \left( \mI - \mB \right)^{-1} = \invJ.
  \end{equation}
  (ii) Let $\gF_{\lambda}(\vh, \vz) = \lambda \gF(\vh, \vz) + (1 - \lambda) \vh$, and
  \begin{equation}
    \frac{\partial \gF_{\lambda}}{\partial \vh} = \lambda \frac{\partial \gF}{\partial \vh} + (1 - \lambda) \mI.
  \end{equation}
  Similar to (\ref{eq:contractive-B}), $\partial \gF_{\lambda} / \partial \vh$ is also a contraction mapping.
  By the Banach Fixed Point Theorem \cite{banach1922operations},
  the sequence $\{ \vh_{t} \}$ converges to an exact fixed point $\vh^{*}$ of $\gF_{\lambda}$,
  which is also a fixed point of $\gF$.

  Denote
  \begin{equation}
    \mU_{t} = \left. \frac{\partial \gF}{\partial \vtheta} \right|_{\vh_{t}},
    \quad
    \mV_{t} = \left. \lambda \frac{\partial \gF}{\partial \vh} \right|_{\vh_{t}} + \left( 1 - \lambda \right) \mI.
  \end{equation}
  Since the function $\gF$ is continuously differentiable \wrt both $\vh$ and $\vtheta$,
  we have
  \begin{equation}
    \lim_{t \to \infty} \mU_{t} = \left. \frac{\partial \gF}{\partial \vtheta} \right|_{\vh^{*}} = \mU_{\infty},
    \quad
    \lim_{t \to \infty} \mV_{t} = \left. \lambda \frac{\partial \gF}{\partial \vh} \right|_{\vh^{*}} + \left( 1 - \lambda \right) \mI = \mV_{\infty}.
  \end{equation}
  According to the conclusion in (i), we have
  \begin{equation}
    \left. \frac{\partial\gF}{\partial\vtheta} \right|_{\vh^*} \left( \mI - \left. \frac{\partial \mathcal{F}}{\partial \vh} \right|_{\vh^{*}} \right)^{-1}
    = \lambda \, \mU_{\infty} \sum_{t = 0}^{\infty} \mV_{\infty}^{t}.
    \label{eq:limit-series-supp}
  \end{equation}
  Comparing \cref{eq:damped-bptt} with \cref{eq:limit-series-supp}, we have
  \begin{equation}
    \begin{split}
      &\, \left\| \frac{\partial \vh_{T}}{\partial \vtheta} - \left. \frac{\partial\gF}{\partial\vtheta} \right|_{\vh^*} \left( \mI - \left. \frac{\partial \mathcal{F}}{\partial \vh} \right|_{\vh^{*}} \right)^{-1} \right\|
      = \lambda \left\| \sum_{t=0}^{T-1} \mU_{t} \prod_{s=t+1}^{T-1} \mV_{s} - \mU_{\infty} \sum_{t = 0}^{\infty} \mV_{\infty}^{t} \right\|
      \\
      \le&\,\lambda \left(
      \underbrace{\left\| \sum_{t=0}^{T-2} \mU_{t} \left( \prod_{s=t+1}^{T-1} \mV_{s} - \mV_{\infty}^{T-t-1} \right) \right\|}_{\bm{\Delta}_{1}}
      + \underbrace{\left\| \sum_{t=0}^{T-1} \left( \mU_{t} - \mU_{\infty} \right) \mV_{\infty}^{T-t-1}  \right\|}_{\bm{\Delta}_{2}}
      + \underbrace{\left\| \mU_{\infty} \sum_{t = T}^{\infty} \mV_{\infty}^{t}  \right\|}_{\bm{\Delta}_{3}}
      \right).
    \end{split}
  \end{equation}
  In the following context, we prove \cref{eq:converge-unroll}
  by showing that $\bm{\Delta}_{1}$, $\bm{\Delta}_{2}$, and $\bm{\Delta}_{3}$ can
  be arbitrarily small when $T$ is sufficiently large.
  \paragraph{Preparations.}
  For any $\epsilon > 0$, since $\mU_{t} \to \mU_{\infty}$ and
  $\mV_{t} \to \mV_{\infty}$ as $t \to \infty$,
  there exists $N \in \mathbb{N}_{+}$ s.t.
  \begin{equation}
    \left\| \mU_{t} - \mU_{\infty} \right\| < \epsilon,
    \quad
    \left\| \mV_{t} - \mV_{\infty} \right\| < \epsilon,
    \quad
    \forall t > N.
  \end{equation}
  Since $\partial \gF_{\lambda} / \partial \vh$ is a contraction mapping,
  there exists $\gamma \in (0, 1)$ s.t.
  \begin{equation}
    \left\| \mV_{t} \right\| \le \gamma,
    \quad
    \left\| \mV_{\infty} \right\| \le \gamma.
  \end{equation}
  Besides, since $\partial \gF / \partial \vtheta$ is a continuous function and $\{ \vh_{t} \}$ is a convergent sequence,
  it follows that $\{ \vh_{t} \}$ is contained by a compact set and that
  $\partial \gF / \partial \vtheta$ is bounded on $\{ \vh_{t} \}$.
  Therefore, there exists $M > 0$, s.t.
  \begin{equation}
    \left\| \mU_{t} \right\| \le M,
    \quad
    t = 0,1,2,\cdots.
  \end{equation}
  Taking $t \to \infty$, we have $\left\| \mU_{\infty} \right\| \le M$.
  \paragraph{For $\bm{\Delta}_{1}$.}
  For $t > N$, consider
  \begin{equation}
    \begin{split}
      &\quad\,\left\| \mU_{t} \left( \prod_{s=t+1}^{T-1} \mV_{s} - \mV_{\infty}^{T-t-1} \right) \right\|
      \\
      &\le \left\| \mU_{t} \right\| \sum_{s=t+1}^{T-1} \left\| \mV_{t+1} \mV_{t+2} \cdots \mV_{s} \mV_{\infty}^{T-s-1} - \mV_{t+1} \mV_{t+2} \cdots \mV_{s-1} \mV_{\infty}^{T-s} \right\|
      \\
      &\le \left\| \mU_{t} \right\| \sum_{s=t+1}^{T-1} \left\| \mV_{t+1} \right\| \left\| \mV_{t+2} \right\| \cdots \left\| \mV_{s-1} \right\| \left\| \mV_{s} - \mV_{\infty} \right\| \left\| \mV_{\infty} \right\|^{T-s-1}
      \\
      &\le M (T - t - 1) \gamma^{T - t - 2} \epsilon,
    \end{split}
  \end{equation}
  and for $t \le N$, we simply have
  \begin{equation}
    \left\| \mU_{t} \left( \prod_{s=t+1}^{T-1} \mV_{s} - \mV_{\infty}^{T-t-1} \right) \right\|
    \le \left\| \mU_{t} \right\| \left( \prod_{s=t+1}^{T-1} \left\| \mV_{s} \right\| + \left\| \mV_{\infty} \right\|^{T-t-1} \right)
    \le 2M \gamma^{T-t-1}.
  \end{equation}
  Therefore, when $T >N + 2$, $\bm{\Delta}_{1}$ can be bounded as follows:
  \begin{equation}
    \begin{split}
      \bm{\Delta}_{1}
      &\le \left( \sum_{t=0}^{N} + \sum_{t=N+1}^{T-2} \right) \left\| \mU_{t} \left( \prod_{s=t+1}^{T-1} \mV_{s} - \mV_{\infty}^{T-t-1} \right) \right\|
      \\
      &\le 2M \sum_{t=0}^{N} \gamma^{T-t-1}
      + M \epsilon \sum_{t=N+1}^{T-2} (T - t - 1) \gamma^{T-t-2}
      \\
      &\le 2M \gamma^{T-N-1} \frac{1 - \gamma^{N+1}}{1 - \gamma}
      + \left( \frac{1 - \gamma^{T-N-2}}{(1 - \gamma)^{2}} - \frac{(T-N-2) \gamma^{T-N-2}}{1 - \gamma} \right) M \epsilon
      \\
      &\le \frac{2M}{1 - \gamma} \gamma^{T-N-1} + \frac{M}{(1 - \gamma)^{2}} \epsilon.
    \end{split}
  \end{equation}
  Since $M / (1 - \gamma)^{2}$ is a constant and $\gamma^{T-N-1} \to 0$ as $T \to \infty$,
  $\bm{\Delta}_{1}$ can be arbitrarily small for a sufficiently large $T$.
  \paragraph{For $\bm{\Delta}_{2}$.}
  Consider
  \begin{equation}
    \left\| \left( \mU_{t} - \mU_{\infty} \right) \mV_{\infty}^{T-t-1} \right\|
    \le \left\| \mU_{t} - \mU_{\infty} \right\| \left\| \mV_{\infty} \right\|^{T-t-1}
    \le
    \begin{cases}
      \gamma^{T-t-1} \epsilon, & \mbox{when}~t \ge N; \\
      2M \gamma^{T-t-1}  & \mbox{when}~t < N.
    \end{cases}
  \end{equation}
  Therefore, when $T >N + 2$, $\bm{\Delta}_{2}$ can be bounded as follows:
  \begin{equation}
    \begin{split}
      \bm{\Delta}_{2}
      &\le \left( \sum_{t=0}^{N} + \sum_{t=N+1}^{T-1} \right) \left\| \left( \mU_{t} - \mU_{\infty} \right) \mV_{\infty}^{T-t-1} \right\|
      \le 2M \sum_{t=0}^{N} \gamma^{T-t-1} + \epsilon \sum_{t=N+1}^{T-1} \gamma^{T-t-1}
      \\
      &\le \frac{2M}{1 - \gamma} \gamma^{T - N - 1} + \frac{\epsilon}{1 - \gamma}.
    \end{split}
  \end{equation}
  Since $1 / (1 - \gamma)$ is a constant and $\gamma^{T-N-1} \to 0$ as $T \to \infty$,
  $\bm{\Delta}_{2}$ can be arbitrarily small for a sufficiently large $T$.
  \paragraph{For $\bm{\Delta}_{3}$.}
  As $t \to \infty$, we have
  \begin{equation}
    \left\| \mU_{\infty} \sum_{t = T}^{\infty} \mV_{\infty}^{t}  \right\|
    \le \left\| \mU_{\infty} \right\| \left\| \mV_{\infty} \right\|^{T} \left\| \left( \mI - \mV_{\infty} \right)^{-1} \right\|
    \le M \cdot \gamma^{T} \cdot \frac{1}{1 - \gamma}
    \to 0.
    \quad
  \end{equation}
  As a result, we obtain the conclusion in \cref{eq:converge-unroll}.
\end{proof}

\subsection{Proof of \cref{thm:convergence-SGD}}
\label{sec:proof-thm3}
\begin{proof}
  Let $\pgn$ be the phantom gradient at the $n^{\text{th}}$ iteration.
  By $\ell$-smoothness of $\gR$, we have
  \begin{equation}
    \begin{split}
      \gR(\vtheta_{n+1})
      &\le \gR(\vtheta_{n}) + \left\langle \nabla \gR(\vtheta_{n}), \vtheta_{n+1} - \vtheta_{n} \right\rangle + \frac{\ell}{2} \left\| \vtheta_{n+1} - \vtheta_{n} \right\|^{2}
      \\
      &= \gR(\vtheta_{n}) - \eta_{n} \left\langle \nabla \gR(\vtheta_{n}), \pgn \right\rangle + \frac{\ell \eta_{n}^{2}}{2} \left\| \pgn \right\|^{2}.
    \end{split}
  \end{equation}
  Let
  \begin{equation}
    \ve_{n} = \left. \frac{\partial \gL}{\partial \vtheta} \right|_{\vtheta = \vtheta_{n}} - \pgn
  \end{equation}
  be the approximation error at the $n^{\text{th}}$ iteration.
  Taking expectation \wrt the first $n$ iterations, we have
  \begin{equation}
    \E_{1 \sim n} \left[ \gR(\vtheta_{n+1}) \right]
    = \E_{1 \sim n-1} \left[ \E_{n} \left[ \gR(\vtheta_{n+1}) \, | \, 1 \sim n-1 \right] \right]
    = \E_{1 \sim n-1} \left[ \E_{n} \left[ \gR(\vtheta_{n+1}) \, | \, \vtheta_{n} \right] \right],
    \label{eq:total-exp}
  \end{equation}
  where the first equality comes from the \textit{law of total expectation},
  while the second from the fact that the stochasticity of the first $n - 1$ steps
  is totally captured by the value $\vtheta_{n}$.
  Consider the inner expectation in \cref{eq:total-exp}, and we omit the condition on
  $\vtheta_{n}$ when no ambiguity is made.
  Note that in the following derivation, all expectations, variances, and covariances are conditioned on $\vtheta_{n}$.
  \begin{equation}
    \begin{split}
      \E_{n} \left[ \gR(\vtheta_{n+1}) \right]
      &\le \E_{n} \left[ \gR(\vtheta_{n}) - \eta_{n} \left\langle \nabla \gR(\vtheta_{n}), \pgn \right\rangle + \frac{\ell \eta_{n}^{2}}{2} \left\| \pgn \right\|^{2}\right]
      \\
      &= \gR(\vtheta_{n}) - \eta_{n} \left\langle \nabla \gR(\vtheta_{n}), \E_{n} \left[ \pgn \right] \right\rangle + \frac{\ell \eta_{n}^{2}}{2} \E_{n} \left[ \left\| \pgn \right\|^{2} \right],
    \end{split}
    \label{eq:R-expansion}
  \end{equation}
  where
  \begin{equation}
    \E_{n} \left[ \pgn \right]
    = \E_{n} \left[ \left. \frac{\partial \gL}{\partial \vtheta} \right|_{\vtheta = \vtheta_{n}} - \ve_{n} \right]
    = \nabla \gR(\vtheta_{n}) - \E_{n} \left[ \ve_{n} \right],
    \label{eq:exp-of-phantom}
  \end{equation}
  and
  \begin{equation}
    \E_{n} \left[ \left\| \pgn \right\|^{2} \right]
    = \left\| \E_{n} \left[ \pgn \right] \right\|^{2} + {\rm tr} \left( {\rm Cov}_{n} \left( \pgn \right) \right).
    \label{eq:var-of-phantom}
  \end{equation}
  Suppose $\| \nabla \gR(\vtheta_{n}) \| \le G$ almost surely,
  and then we have
  \begin{equation}
    \left\| \E_{n} \left[ \pgn \right] \right\|^{2}
    = \left\| \nabla \gR(\vtheta_{n}) - \E_{n} \left[ \ve_{n} \right] \right\|^{2}
    \le (G + \epsilon)^{2}.
    \label{eq:bound-phantom-norm}
  \end{equation}
  Moreover, by the properties of covariance,
  \begin{equation}
    \begin{split}
      &\;{\rm tr} \left( {\rm Cov}_{n} \left( \pgn \right) \right)
      = {\rm tr} \left( {\rm Cov}_{n} \left( \left. \frac{\partial \gL}{\partial \vtheta} \right|_{\vtheta = \vtheta_{n}} - \ve_{n} \right) \right)
      \\
      =&\; {\rm tr} \left( {\rm Cov}_{n} \left( \left. \frac{\partial \gL}{\partial \vtheta} \right|_{\vtheta = \vtheta_{n}} \right) \right)
      + {\rm tr} \left( {\rm Cov}_{n} \left( \ve_{n} \right) \right)
      - 2 \, {\rm tr} \left( {\rm Cov}_{n} \left( \left. \frac{\partial \gL}{\partial \vtheta} \right|_{\vtheta = \vtheta_{n}}, \ve_{n} \right) \right)
      \\
      \le&\; 2 \, {\rm tr} \left( {\rm Cov}_{n} \left( \left. \frac{\partial \gL}{\partial \vtheta} \right|_{\vtheta = \vtheta_{n}} \right) \right)
      + 2 \, {\rm tr} \left( {\rm Cov}_{n} \left( \ve_{n} \right) \right),
    \end{split}
    \label{eq:bound-phantom-cov}
  \end{equation}
  where the last inequility comes from
  \begin{equation}
    \begin{split}
      \left| {\rm tr} \left( {\rm Cov} \left( \va, \vb \right) \right) \right|
      &\le \sum_{i} \left| {\rm Cov} \left( a_{i}, b_{i} \right) \right|
      \le \sum_{i} \sqrt{{\rm Var} \left( a_{i} \right) {\rm Var} \left( b_{i} \right)}
      \le \sum_{i} \frac{{\rm Var} \left( a_{i} \right) + {\rm Var} \left( b_{i} \right)}{2}
      \\
      &= \frac{1}{2} \left( {\rm tr} \left( {\rm Cov} \left( \va \right) \right) + {\rm tr} \left( {\rm Cov} \left( \vb \right) \right) \right).
    \end{split}
  \end{equation}
  By the Popoviciu's inequality on variances \cite{popoviciu1935equations},
  the second term in (\ref{eq:bound-phantom-cov}) can be bounded by $d_{\vtheta} \epsilon^{2}$, \ie
  \begin{equation}
    {\rm tr} \left( {\rm Cov}_{n} \left( \ve_{n} \right) \right)
    \le d_{\vtheta} \epsilon^{2},
  \end{equation}
  where $d_{\vtheta}$ denotes the dimension of $\vtheta$.
  Finally, since the gradient estimator $\partial \gL / \partial \vtheta$ has
  a bounded covariance, there exists $M > 0$, s.t.
  \begin{equation}
    {\rm tr} \left( {\rm Cov}_{n} \left( \left. \frac{\partial \gL}{\partial \vtheta} \right|_{\vtheta = \vtheta_{n}} \right) \right) \le M,
    \quad
    \mbox{almost surely}.
    \label{eq:bound-cov}
  \end{equation}
  Combining (\ref{eq:R-expansion}), (\ref{eq:exp-of-phantom}),
  (\ref{eq:var-of-phantom}), (\ref{eq:bound-phantom-norm}),
  (\ref{eq:bound-phantom-cov}), and (\ref{eq:bound-cov}),
  we have
  \begin{equation}
    \begin{split}
      \E_{n} \left[ \gR(\vtheta_{n+1}) \right]
      &\le \gR(\vtheta_{n}) - \eta_{n} \left\| \nabla \gR(\vtheta_{n}) \right\|^{2} + \eta_{n} \left\langle \nabla \gR(\vtheta_{n}), \E_{n} \left[ \ve_{n} \right] \right\rangle + K \eta_{n}^{2},
      \\
      &\le \gR(\vtheta_{n}) - \eta_{n} \left\| \nabla \gR(\vtheta_{n}) \right\|^{2} + \eta_{n} \left\| \nabla \gR(\vtheta_{n}) \right\| \left\| \E_{n} \left[ \ve_{n} \right] \right\| + K \eta_{n}^{2}
      \\
      &\le \gR(\vtheta_{n}) - \eta_{n} \left\| \nabla \gR(\vtheta_{n}) \right\|^{2} + \eta_{n} G \epsilon + K \eta_{n}^{2},
    \end{split}
    \label{eq:bound-inner}
  \end{equation}
  where $K = \ell \left( (G + \epsilon)^{2} + 2M + 2 d_{\vtheta} \epsilon^{2} \right) / 2$ is a constant.
  Substitute (\ref{eq:bound-inner}) into \cref{eq:total-exp}, and it becomes
  \begin{equation}
    \E_{1 \sim n} \left[ \gR(\vtheta_{n+1}) \right]
    \le \E_{1 \sim n-1} \left[ \gR(\vtheta_{n}) \right] - \eta_{n} \E_{1 \sim n-1} \left[ \left\| \nabla \gR(\vtheta_{n}) \right\|^{2} \right] + \eta_{n} G \epsilon + K \eta_{n}^{2}.
  \end{equation}
  By taking a summation over the first $N$ steps, we have
  \begin{equation}
    \begin{split}
      \E_{1 \sim N} \left[ \sum_{n=1}^{N} \eta_{n} \left\| \nabla \gR(\vtheta_{n}) \right\|^{2} \right]
      &\le \gR(\vtheta_{1}) - \E_{1 \sim N} \left[ \gR(\vtheta_{N+1}) \right] + G \epsilon \sum_{n=1}^{N} \eta_{n} + K \sum_{n=1}^{N} \eta_{n}^{2}
      \\
      &\le \gR(\vtheta_{1}) - m + G \epsilon \sum_{n=1}^{N} \eta_{n} + K \sum_{n=1}^{N} \eta_{n}^{2},
    \end{split}
  \end{equation}
  where $m = \inf_{\vtheta} \gR(\vtheta)$ since $\gR$ is lower-bounded.
  Dividing a factor of $\sum_{n=1}^{N} \eta_{n}$, we have
  \begin{equation}
    \E_{1 \sim N} \left[ \frac{\sum_{n=1}^{N} \eta_{n} \| \nabla \gR(\vtheta_{n}) \|^2}{\sum_{n=1}^{N} \eta_{n}} \right]
    \le G \epsilon + \frac{\gR(\vtheta_{1}) - m}{\sum_{n=1}^{N} \eta_{n}} + K \frac{\sum_{n=1}^{N} \eta_{n}^{2}}{\sum_{n=1}^{N} \eta_{n}}.
    \label{eq:bound-objective}
  \end{equation}
  Since $\eta_{n} = \gO (1 / \sqrt{n})$, it follows that
  \begin{equation}
    \sum_{n=1}^{N} \eta_{n} = \gO \left( \sqrt{N} \right),
    \quad
    \frac{\sum_{n=1}^{N} \eta_{n}^{2}}{\sum_{n=1}^{N} \eta_{n}} = \gO \left( \frac{\log{N}}{\sqrt{N}} \right).
    \label{eq:O-N}
  \end{equation}
  Combining (\ref{eq:bound-objective}) and \cref{eq:O-N} concludes the proof.
\end{proof}

\paragraph{Remark 3.}
The assumption that $\gR$ has almost-surely bounded gradient
at $\{ \vtheta_{n} \}_{n = 0}^{N}$ is reasonable.
Because of the existence of norm-based regularizations, \eg weight decay, we can assume
$\vtheta$ is almost surely optimized within a compact set in the parameter space.
If we further assume $\gR$ is continuously differentiable,
the almost-sure boundedness of $\left\| \nabla \gR \right\|$ within the compact set
follows its continuity.

\paragraph{Remark 4.}
We justify the assumption that the gradient in \cref{eq:true-grad} has a bounded covariance.
For the SGD algorithm, the stochasticity of the gradient in \cref{eq:true-grad} comes from the random sampling
of the training example (or the training mini-batch) from the dataset.
Since there are finite samples in the training set, the covariance of \cref{eq:true-grad} remains finite.
Moreover, as \cref{thm:convergence-of-phantom} only considers a finite training schedule,
\ie $N$ steps, the possible combination of the selected sample (or mini-batch) at each step is still finite
(even though its number grows combinatorially).
Therefore, it is reasonable to assume the gradient in \cref{eq:true-grad} has a bounded covariance.

\section{Experiment Details}
\label{sec:experiment-supp}
In this section, we introduce the experimental settings of this paper in detail
and discuss some additional findings of training implicit models.

\subsection{Synthetic Setting}
For the synthetic setting, the following model is used:
\begin{equation}
    \vh^{*} = \gF \left(\vh^* + \vu\right)
\end{equation}
where $\gF$ is an 1-layer network with spectral normalization \cite{takeru2018spectral}, and $\vu,\,\vh^* \in \R^{N \times D}$.
The loss $\gL$ is given by the mean squared error (MSE) between $\vh^*$ and $\vy$.
We choose $N=32$ and $D=128$ and randomly sample $50000$ data pairs $\left(\vu,\,\vy \right)$ to compute the gradient $\partial \gL / \partial \vu$.
 
\begin{figure}[!t]
    \centering
    \begin{overpic}[width=\linewidth]{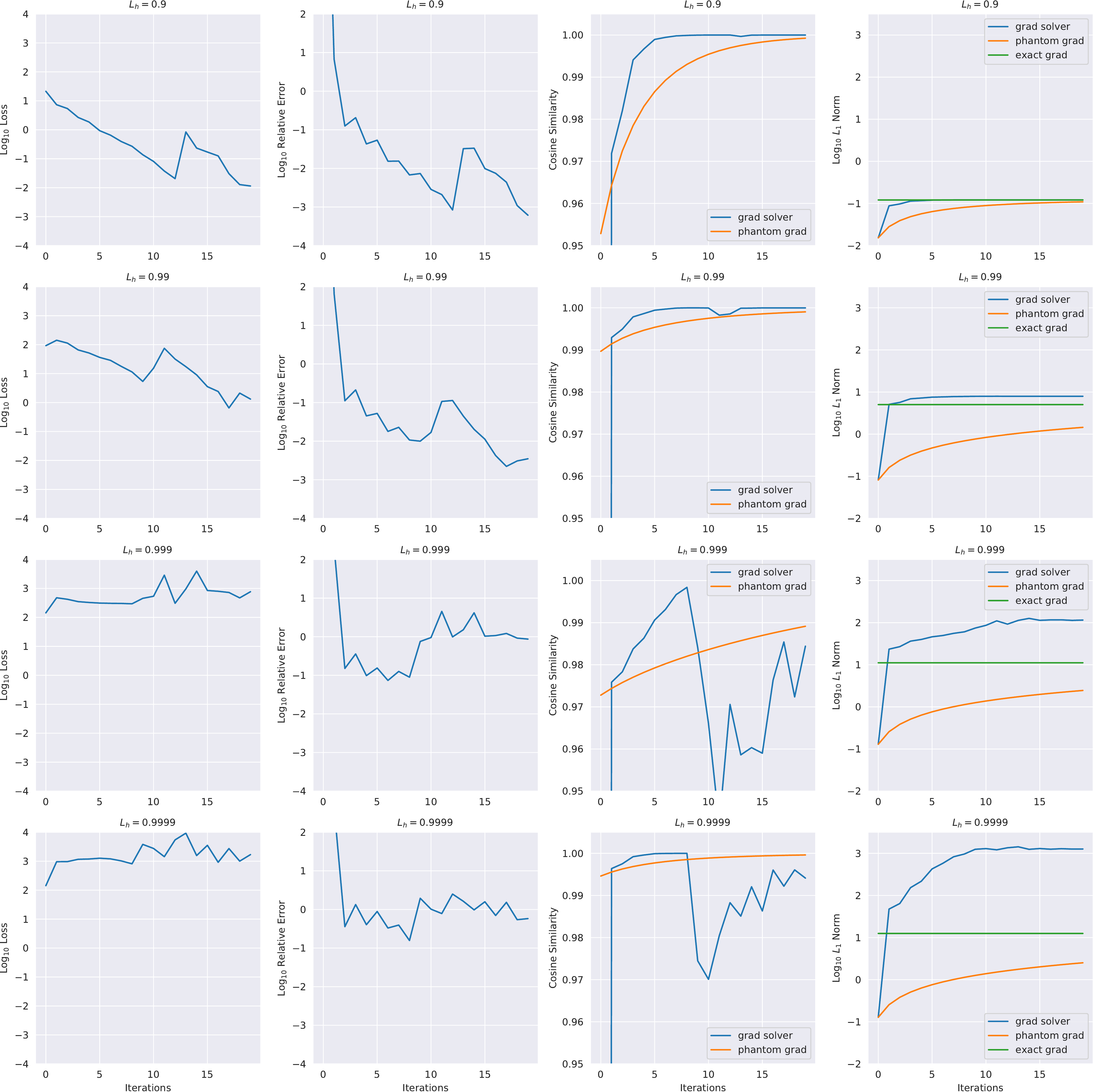}
    \end{overpic}
    \caption{Visualization of gradient solvers under different $L_\vh$.}
    \label{fig:optimization-supp}
\end{figure}

We generate a symmetric weight matrix for the network and constrain the Lipschitz constant $L_\vh$ to a given level using spectral normalization.
For the visualization in the main paper, we adopt $L_\vh = 0.9$. For the additional visualization on the stability of the solver in \cref{fig:optimization-supp}, we choose $L_\vh$ from $\left\{0.9,\,0.99,\,0.999,\,0.9999\right\}$.

To solve $\vh^*$, we employ the fixed-point iteration as the solver.
For the synthetic setting, we use 100 fixed-point iterations to obtain $\vh^*$ that satisfies the relative error $\| \vh^* - \gF(\vh^*, \vu)\| / \| \vh^* \| \le {10}^{-5}$.
For the visualization in \cref{fig:optimization-supp}, we also apply 100 fixed-point iterations for each $L_\vh$.

\subsection{Ablation Setting}

For the ablation setting, we use the original MDEQ-Tiny \cite{bai2020multiscale} model (170K parameters) on CIFAR-10 \cite{CIFAR} classification without any architecture modification.
Therefore, the performance gain upon the \sota method is due to
the improved training efficiency thanks to the proposed phantom gradient.

The experiments are conducted without data augmentation as in \cite{bai2020multiscale}.
The training schedule, batch size, cosine learning rate annealing strategy, and other hyperparameters are kept unchanged for all ablation experiments.
We also follow the official training protocol of MDEQ\footnote{Code available at \url{https://github.com/locuslab/mdeq}.} to reproduce its result.

For the training protocol without pretraining, we substitute the unrolled pretraining stage by implicit differentiation.
For the training protocol without Dropout, we remove the variational Dropout from the model.
We also experiment with the SGD optimizer under the standard hyperparameter setting,
\ie a learning rate of $0.1$, a momentum of $0.9$, and a weight decay of $0.0001$.

We train the MDEQ model using the two types of phantom gradient
with the SGD optimizer (under the hyperparameters mentioned above) and other hyperparameters unchanged from the original setting.
The model is trained without shallow-layer pretraining, suggesting an $\gO(k)$ and
$\gO(1)$ peak memory usage for the unrolling-based and the Neumann-series-based phantom gradient, respectively.
In both cases, the damped fixed-point iteration starts at the solution obtained by the Broyden's method.

We monitor the Jacobian spectral radius $\rho(\partial \gF / \partial \vh)$ during training for both forms of phantom gradient.
It shows that the radius can grow without restriction for the state-free NPG
when the phantom gradient includes high-order terms and cannot exactly
match the gradient of a computational sub-graph.
A similar phenomenon is observed when using the state-free gradient estimate
from implicit differentiation with considerable numerical errors in the forward and backward passes \cite{bai2021stabilizing}. 
On the contrary, for the state-dependent UPG, the Jacobian spectral radius is kept within a reasonable region during training thanks to the implicit Jacobian regularization.

\subsection{Experiments at Scale}

For large-scale experiments, we adopt MDEQ and MDEQ-Small on the CIFAR-10 \cite{CIFAR} and ImageNet \cite{ILSVRC15} benchmarks, respectively, DEQ (PostLN) \cite{bai2019deep} and DEQ (PreLN) \cite{bai2021stabilizing} on the Wikitext-103 \cite{WIKI} dataset, and IGNN \cite{gu2020implicit} on graph classification (COX2, PROTEINS) and node classification (PPI) benchmarks.

\paragraph{ConvNet-based Implicit Models on Vision Datasets.}
To train MDEQ on CIFAR-10, we employ the UPG with $\lambda = 0.5$ and $k=5$, \ie $\mA_{5,0.5}$.
Besides, we use the SGD optimizer with a learning rate of $0.1$, a momentum of $0.9$,
and a weight decay of $0.0001$, and keep other experimental settings unchanged,
including the number of training epochs, the batch size, the learning rate annealing strategy, \etc

We adopt two settings on ImageNet. 
The first setting follows the practice of \cite{bai2020multiscale} to pretrain the model for the same number of epochs.
Afterwards, the UPG with $\mA_{5,0.6}$ is used to train the model for the remaining training schedule.
This setting achieves a test accuracy of $75.2\%$.
In the second setting, we adopt the UPG to train the implicit model throughout,
leaving the UPG to automatically transit from the pretraining stage to the regular training stage.
This setting demonstrates a test accuracy of $75.7\%$.
The difference confirms that the automatic transition property of UPG
helps alleviate the burden of hyperparameter tuning, \ie the number of steps
in the pretraining stage, and benefit to the final performance as well.

We also verify the implicit Jacobian regularization from the UPG on ImageNet.
By calculating the Jacobian spectral radius of the trained model on the validation set through the power method,
we find that the radius $\rho(\partial \gF_{\lambda} / \partial \vh)$ is retained around $1$,
although the radius of the equilibrium module $\rho(\partial \gF / \partial \vh)$ usually exceeds $1$ (but also remains bounded).
This finding provides us with a potential path to explain why the damping operation
can enhance the naive unrolling to match or even surpass the standard implicit differentiation for ConvNets on vision tasks.
We conjecture that the damping operation allows the equilibrium module to evolve
within a wider range, \eg $\rho(\partial \gF / \partial \vh) > 1$,
which contributes to its better representative capacity,
while maintaining stability regarding the backward pass, \ie $\rho(\partial \gF_{\lambda} / \partial \vh) \approx 1$.

\paragraph{Transformer-based Implicit Models on Language Datasets.}
For language modeling on Wikitext-103, we follow the official training protocol
of the DEQ model \cite{bai2019deep}.
However, the UPG leads to inferior generalization capacity on the test set
while the training loss is similar to that of implicit differentiation. 
The NPG even fails to optimize the DEQ (PostLN) model
unless the explicit Jacobian regularization \cite{bai2021stabilizing} or
the adaptive damping factor, \eg $\lambda = 1 / \rho(\partial \gF / \partial \vh)$,
is applied.

The performance discrepancy of different implicit models suggests the following perspective.
The loss landscape and training strategy are the two sides of the same coin.
Architecture, dataset, and loss function jointly define the loss landscape that has considerable impact on the preferable training strategy.
For the ConvNet-based implicit model trained on vision tasks,
the loss landscape is likely more regular
so that the model trained on the phantom gradient can extricate itself from severe overfitting and achieve remarkable performance with acceleration despite the biased gradient estimate (which means the approximation error cannot be easily zeroed out by taking the expectation over the data distribution).
For the Transformer-based implicit model on language processing tasks,
in contrast, it is more arduous to employ the phantom gradient
due to a lack of regularity of the loss landscape,
thus inspiring us to supplement with additional regularization on the loss landscape.

To this end, we introduce the explicit Jacobian regularization (JR) \cite{bai2021stabilizing}
to strengthen the regularity of the loss landscape.
The training protocol follows the official source of DEQ with JR\footnote{Code available at \url{https://github.com/locuslab/deq}.}.
Note that with the implicit Jacobian regularization effect of the UPG,
the weight of the explicit JR can be significantly reduced, \eg from $2.0$ to $0.1$,
and the training stability is still maintained.
Meanwhile, the explicit JR can also play a vital role in alleviating overfitting for the UPG.
Combining the UPG with explicit JR demonstrates an impressive test perplexity of $24.4$ with $2.2 \times$ training acceleration (with $14$ forward Broyden iterations),
and a test perplexity of $24.0$ with $1.7 \times$ training acceleration (with $20$ forward Broyden iterations).

Our results indicate that it is more tactful to understand the training strategy
combined with the loss landscape instead of only focusing on the former but neglecting the latter.

\paragraph{GNN-based Implicit Models on Graph Datasets.}
To conduct experiments on graph datasets,
we follow the default architectures and training settings of the IGNN model \cite{gu2020implicit}\footnote{
  Code available at \url{https://github.com/SwiftieH/IGNN}.
}.
We employ different damping factor $\lambda$ for both graph classification and node classification.
In the experiments, we encounter the training stability issue for IGNN on the PPI node classification task.
Specifically, the IGNN model suffers from training collapse
when using either the UPG or the exact gradient by implicit differentiation.
Hence the best result from three runs is reported for this task.
We conjecture that the stability issue comes from hyperparameter selection 
regarding the projected gradient in IGNN,
since it is not easy to figure out the proper hyperparameters for well-posedness.
For graph classification, the stability issue is not observed.

\subsection{Additional Analysis on the Gradient Solver}

To illustrate the vulnerability the gradient solver for implicit differentiation in the ill-conditioned cases,
we provide the optimization dynamics in \cref{fig:optimization-supp} and its comparison with the phantom gradient in the synthetic setting.
We plot (1) the optimization objective $\| (I - \partial \gF / \partial \vh) \hat{\vg} - \partial \gL / \partial \vh \|$,
(2) the relative error $\| (I - \partial \gF / \partial \vh) \hat{\vg} - \partial \gL / \partial \vh \| / \| \hat{\vg} \|$,
(3) the cosine similarity between the solved gradient $\hat{\vg}$ (or the phantom gradient) and the exact gradient $\vg$,
and (4) the $L_1$ norm of the solved gradient $\hat{\vg}$, the phantom gradient, and the exact gradient $\vg$.
Here, in the context of optimization, $\hat{\vg}$ is the solution of the backward linear system solved by the Broyden's method.

\cref{fig:optimization-supp} shows that the gradient solver diverges in ill-conditioned situations.
It is shown that the phantom gradient demonstrates much better stability,
especially in the extremely ill-conditioned cases, \eg $L_{\vh} = 0.9999$.
As for the Broyden's method, more optimization steps do not necessarily make
the solved gradient more aligned to the exact gradient, as indicated by the oscillating cosine similarity.
Besides, the norm of the solved gradient also tends to explode in the optimization process,
while the phantom gradient maintains a moderate norm throughout.

\end{document}